\documentclass[runningheads]{llncs}

\usepackage{eccv}

\usepackage{eccvabbrv}

\usepackage{graphicx}
\usepackage{booktabs}

\usepackage[accsupp]{axessibility}  %

\usepackage[most]{tcolorbox}
\usepackage{booktabs}
\usepackage{tabularx}
\usepackage{url}
\usepackage{adjustbox}
\usepackage{dirtytalk}
\usepackage{soul, color}
\usepackage{booktabs}
\usepackage{multirow}
\usepackage{graphicx}
\usepackage{subcaption}
\usepackage{caption}
\usepackage{tikz}
\usepackage{xspace}
\usepackage{listings}
\usepackage{xcolor}
\usepackage{array}
\usepackage{afterpage}

\usepackage{orcidlink}

\begin{document}

\title{Lost in Translation: Latent Concept Misalignment in Text-to-Image Diffusion Models} 

\titlerunning{Latent Concept Misalignment in Text-to-Image Diffusion Models}

\author{
Juntu Zhao\inst{1}\orcidlink{0009-0003-6638-2283} \and
Junyu Deng\inst{3}\textsuperscript{*}\orcidlink{0009-0004-6313-7378} \and
Yixin Ye\inst{1}\textsuperscript{*}\orcidlink{0009-0006-1882-2960} \and
Chongxuan Li\inst{4}\orcidlink{0000-0002-0912-9076} 
\and
\\
Zhijie Deng\inst{1}\textsuperscript{\dag}\orcidlink{0000-0002-0932-1631} \and
Dequan Wang\inst{1,2}\textsuperscript{\dag}\orcidlink{0000-0001-8270-8448}
}

\authorrunning{J.~Zhao et al.}

\institute{
Shanghai Jiao Tong University \and
Shanghai Artificial Intelligence Laboratory \and
Fudan University \and
Renmin University of China
}

\maketitle

\newcommand{\ourDataset}{LC-Mis\xspace}
\newcommand{\ourMethod}{MoCE\xspace}
\newcommand{\ourProblem}{LC-Mis\xspace}
\newcommand{\ourMetric}{Multi-Concept Disparity\xspace}
\newcommand{\ourScore}{Description-Concept Coordination Score\xspace}

\newcommand{\cx}[1]{\textcolor{green}{\xspace[cx:\xspace#1]\xspace}}
\newcommand{\junyu}[1]{\textcolor{red}{\xspace[\xspace#1]\xspace}}
\newcommand{\zhijie}[1]{\textcolor{red}{\xspace[zhijie:\xspace#1]\xspace}}
\newcommand{\dequan}[1]{\textcolor{orange}{\xspace[dequan:\xspace#1]\xspace}}
\newcommand{\juntu}[1]{\textcolor{blue}
{\xspace[juntu:\xspace#1]\xspace}}
\newcommand{\yixin}[1]{\textcolor{brown}{\xspace[yixin:\xspace#1]\xspace}}
\newcommand{\gpt}{GPT-3.5\xspace}
\newcommand{\llm}{LLM\xspace}
\newcommand{\sdxl}{SDXL\xspace}
\newcommand{\mj}{Midjourney\xspace}
\newcommand{\ti}{text-to-image\xspace}
\newcommand{\clipscore}{Clipscore\xspace}
\newcommand{\imgreward}{Image-Reward\xspace}
\newcommand{\dynamic}{MoCE\xspace}
\newcommand{\distance}{Distance Score\xspace}
\newcommand{\conceptA}{$\mathcal{A}$\xspace} %
\newcommand{\conceptB}{$\mathcal{B}$\xspace} %
\newcommand{\conceptC}{$\mathcal{C}$\xspace} %
\newcommand{\hard}{$\mathcal{D}_h$\xspace} %
\newcommand{\harddyn}{$\overline{\mathcal{D}_h}$\xspace} %
\newcommand{\valuable}{$\mathcal{D}_v$\xspace} %
\newcommand{\simplep}{$\mathcal{D}_s$\xspace} %
\newcommand{\socratic}{Socratic Reasoning\xspace}
\definecolor{darkred}{rgb}{0.5, 0, 0} %
\newcommand{\PositiveFewShot}{\textcolor{feishublueplus}{[positive]}\xspace}
\definecolor{darkblue}{rgb}{0, 0, 0.5} %
\definecolor{qing}{RGB}{0, 255, 255}
\newcommand{\NegativeFewShot}{\textcolor{feishuorangeplus}{[negative]}\xspace}
\definecolor{feishugray}{RGB}{242,243,245}
\definecolor{feishured}{RGB}{252,242,241}
\definecolor{feishuorange}{RGB}{253,246,236}
\definecolor{feishuyellow}{RGB}{254,255,241}
\definecolor{feishugreen}{RGB}{50,200,50}
\definecolor{feishublue}{RGB}{241,244,254}
\definecolor{feishupurple}{RGB}{246,241,253}
\definecolor{feishuorangeplus}{RGB}{253,175,64}
\definecolor{feishublueplus}{RGB}{41,86,165}
\definecolor{feishupurpleplus}{RGB}{81,70,154}
\definecolor{levelfivebrown}{RGB}{163, 84, 83}

\begin{abstract}
  Advancements in text-to-image diffusion models have broadened extensive downstream practical applications, but such models often encounter misalignment issues between text and image. 
  Taking the generation of a combination of two disentangled concepts as an example, say given the prompt \say{a tea cup of iced coke}, existing models usually generate a glass cup of iced coke because the iced coke usually co-occurs with the {glass cup} instead of the tea one during model training. 
  The root of such misalignment is attributed to the confusion in the latent semantic space of text-to-image diffusion models, and hence we refer to the \say{a tea cup of iced coke} phenomenon as Latent Concept Misalignment (LC-Mis). 
  We leverage large language models (LLMs) to thoroughly investigate the scope of LC-Mis, and develop an automated pipeline for aligning the latent semantics of diffusion models to text prompts. %
  Empirical assessments confirm the effectiveness of our approach, substantially reducing LC-Mis errors and enhancing the robustness and versatility of text-to-image diffusion models. 
  Our \href{https://github.com/RossoneriZhao/iced_coke}{\textit{code}} and \href{https://huggingface.co/datasets/Tutu1/LatentConceptMisalignment/tree/main/version1}{\textit{dataset}} have been available online for reference.
  \keywords{Text-to-image diffusion models \and Misalignment \and Large Language Models}
\end{abstract}

\section{Introduction}
\label{sec:intro}
\begin{figure*}[t]
    \begin{center}
        \adjustbox{max width=\linewidth}{
            \begin{tabular}{c c c c}
            \includegraphics[height=3cm]{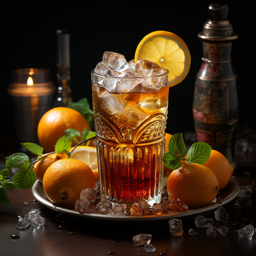} &
            \includegraphics[height=3cm]{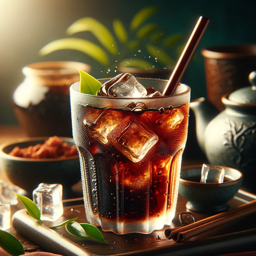} &
            \includegraphics[height=3cm]{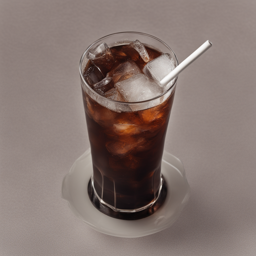} &
            \includegraphics[height=3cm]{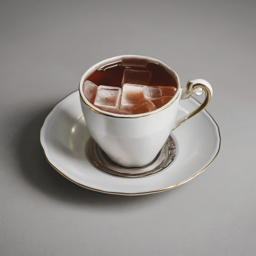} \\
            \textcolor{red}{Midjourney~\cite{midjourney}}  & \textcolor{red}{Dall·E 3~\cite{openai2023dalle3}} & \textcolor{red}{SDXL~\cite{podell2023sdxl}} & \textcolor{feishugreen}{\textbf{MoCE} (ours)}
            \end{tabular}
        }
    \end{center}
    \caption{Teaser figures, from three models and our approach \dynamic, showcase a classic example of Latent Concept Misalignment (\ourProblem) in this study: a tea cup of iced coke. Here, a glass cup, an unfamiliar object, substitutes the anticipated tea cup. We denote the iced coke as Concept \conceptA, the tea cup as Concept \conceptB, and introduce a latent Concept \conceptC—the glass. This combination of \conceptA, \conceptB, and \conceptC forms our investigative focus.
    }
    \label{fig:teaser_coke_cup}
\end{figure*}

\renewcommand{\thefootnote}{\fnsymbol{footnote}}
\footnotetext[1]{Equal contribution. \textsuperscript{\dag} Corresponding authors.}
Text-to-image synthesis~\cite{mansimov2015generating,reed2016generative,zhang2017stackgan,xu2017attngan,li2019controllable,ramesh2021zero,ding2021cogview,wu2022nuwa,yu2022scaling,nichol2021glide,saharia2022photorealistic,rombach2022high,ramesh2022hierarchical,song2023consistency} via diffusion models has made remarkable progress, where high-quality images are generated given text prompts~\cite{rombach2022high, podell2023sdxl}. 
However, a significant limitation of existing models is that they can easily face visual-textual misalignment in practice, where certain elements in the input text are overlooked in generated images. 
As shown in Figure \ref{fig:teaser_coke_cup}, none of {Midjourney~\cite{midjourney}}, {Dall·E 3~\cite{openai2023dalle3}}, and {SDXL~\cite{podell2023sdxl}} can craft an image containing \say{a tea cup of iced coke}.
Instead, these models exhibit a preference for generating a glass cup due to inherent biases in concept combination during the training process of the models.

Referring to the two main concepts (e.g., \say{iced coke} and \say{tea cup}) as \conceptA and \conceptB, and a latent concept that inherently correlates with \conceptA or \conceptB as \conceptC (e.g., \say{glass cup}), by our experience, it is non-trivial to address the misalignment problem based on naive prompt engineering.
We refer to such an inherent issue as 
Latent Concept Misalignment (\ourProblem).
Unlike basic works that only focus on the mutual encroachment of \conceptA and \conceptB~\cite{wang2023compositional, du2023reduce, liu2022compositional, li2023gligen, chefer2023attend}, our problem involves a latent concept \conceptC that has never been mentioned in the text prompt. %
The emergence of this phenomenon can lead to the absence of the expected concept \conceptB (\say{tea cup}) in the generated output.

We devise an efficient pipeline with the aid of Large Language Models~\cite{brown2020language, chatgpt, achiam2023gpt} (LLMs) to discover an extensive set of \ourProblem examples.
Specifically, human researchers first patiently guide the LLM to gradually understand and delve into the logic behind \ourProblem. 
The LLM is then leveraged to generate additional \ourProblem concept pairs according to its own understanding. 
After acquiring a substantial number of \ourProblem concept pairs, state-of-the-art diffusion models, including Midjourney~\cite{midjourney} and SDXL~\cite{podell2023sdxl}, are employed to synthesize high-quality image samples for evaluation.
Subsequently, expert human researchers meticulously evaluate the generated images to identify and select concept pairs that accurately manifest \ourProblem. This rigorous evaluation process culminates in the formation of our dataset.

\begin{figure}[t]
    \centering
    \includegraphics[width=\linewidth]{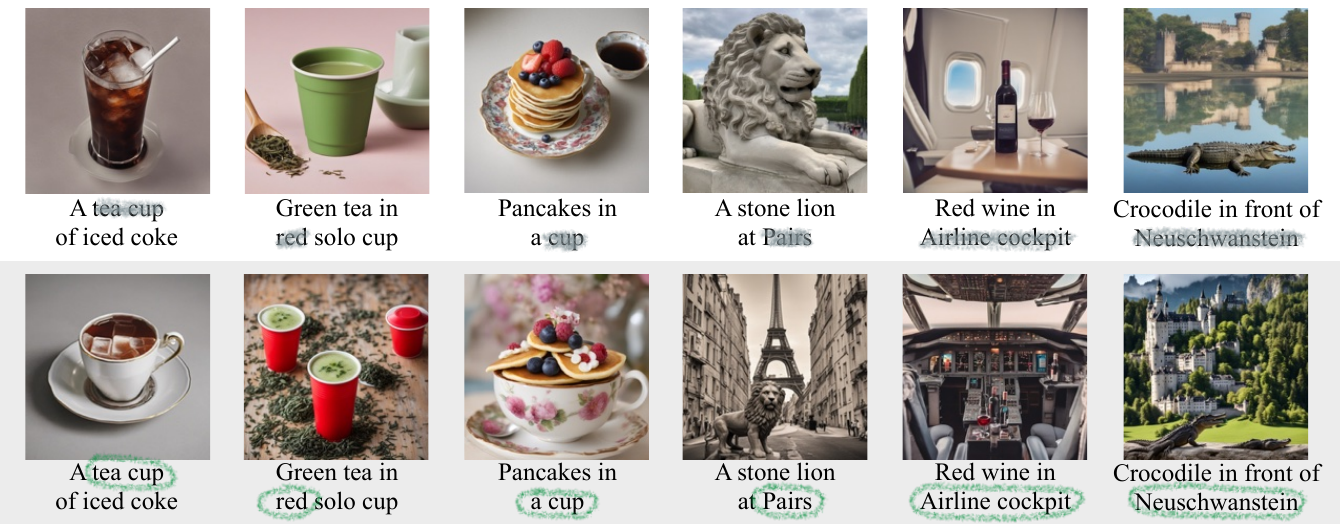}
    \caption{We display the issue of  Latent Concept Misalignment (\ourProblem). In the first row of the images, even the most advanced text-to-image models (SDXL) fail to faithfully generate the specified concepts. We have developed an autonomous pipeline to explore this issue and proposed a hotfix, MoCE, to simply fix it.}
    \label{fig:teaser}
\end{figure}

We introduce Mixture of Concept Experts (\ourMethod) to enhance the alignment between images and text in text-to-image diffusion models and hence mitigate the \ourProblem issue.
Specifically, inspired by the sequential rule of human drawing when faced with a concept pair (\conceptA, \conceptB), we divide the entire sampling process of diffusion models into two phases based on the drawing advice given by an LLM, where only one of \conceptA and \conceptB is provided in the first phase and then the complete text prompt is used in the second phase. 
Once the image synthesis is complete, a quantitative metric based on proven benchmarks, such as \clipscore~\cite{hessel2021clipscore} and \imgreward~\cite{xu2023imagereward}, is employed to measure the alignment between the generated images and the text prompt. This measurement is then used to iteratively adjust the lengths of the two phases mentioned above, employing a binary search method, until the generated images meet the expectation. 
Empirical case studies, grounded in thorough human evaluation, provide confirmation that our approach significantly mitigates the \ourProblem issue. Furthermore, it also enhances both the applicability and flexibility of text-to-image diffusion techniques across diverse fields. This improvement is clearly demonstrated in Figure~\ref{fig:teaser}. The models we used in this paper are those available online as of \textbf{October 1, 2023}. See Section~\ref{sec:new_model} for the discussion on the latest models.

\ 

\noindent
\textbf{Here, we summarize our contributions in the following 2 aspects:}

\begin{itemize}
    \item We investigate the neglected Latent Concept Misalignment (LC-Mis) issue within existing text-to-image diffusion models, introduce an LLM-based pipeline to collect our \ourProblem dataset.

    \ 

    \item We propose to split the concepts in text prompt and input them into different phases of the diffusion model generation process, effectively mitigating the \ourProblem issue.
\end{itemize}

\section{Related Work}
\label{sec:related}
\textbf{Diffusion Models\ }
Diffusion models~\cite{ho2020denoising} predict noise in noisy images and produce high-quality outputs after training. Having gained considerable attention, they are now considered the state-of-the-art in image generation. These models find applications in various domains, such as label-to-image synthesis~\cite{dhariwal2021diffusion, rombach2022high}, text-to-image generation~\cite{ramesh2022hierarchical, rombach2022high, podell2023sdxl}, image editing~\cite{couairon2022diffedit, meng2021sdedit, kawar2023imagic}, and video generation~\cite{ho2022imagen}. Specifically, text-to-image generation is of practical importance. Open-source and commercial solutions like Stable Diffusion~\cite{rombach2022high, podell2023sdxl}, Midjourney~\cite{midjourney} and Dall·E 3~\cite{openai2023dalle3} have achieved success, bolstered by advancements in neural networks~\cite{ronneberger2015u, vaswani2017attention} and large datasets of image-text pairs~\cite{sharma2018conceptual, schuhmann2022laion}. Nevertheless, the question of whether these models truly innovate or merely generate combinations encountered in their training data remains unresolved. In this study, we investigate this issue by examining text-to-image diffusion models in the context of unconventional concept pairings, namely \ourProblem.
 \\
 \\
\noindent
\textbf{Misalignment Issues\ }
While state-of-the-art generative models frequently produce high-quality, realistic images, they struggle with certain concept combinations. Such models usually mimic combinations seen in training data. Prior work has emphasized spatial conflicts where multiple entities coexist in close proximity~\cite{wang2023compositional, du2023reduce, liu2022compositional, li2023gligen, chefer2023attend}. Distinct from these investigations, our focus lies on  Latent Concept Misalignment, illustrated by phrases such as \say{a tea cup of iced coke.} Through rigorous experimentation, we investigate this challenge and introduce a benchmark alongside a hotfix solution.

\section{Benchmark: Collecting Data on Latent Concept Misalignment (\ourDataset)}
\label{sec:dataset}
\begin{figure}[t]
    \centering
    \includegraphics[width=\linewidth]{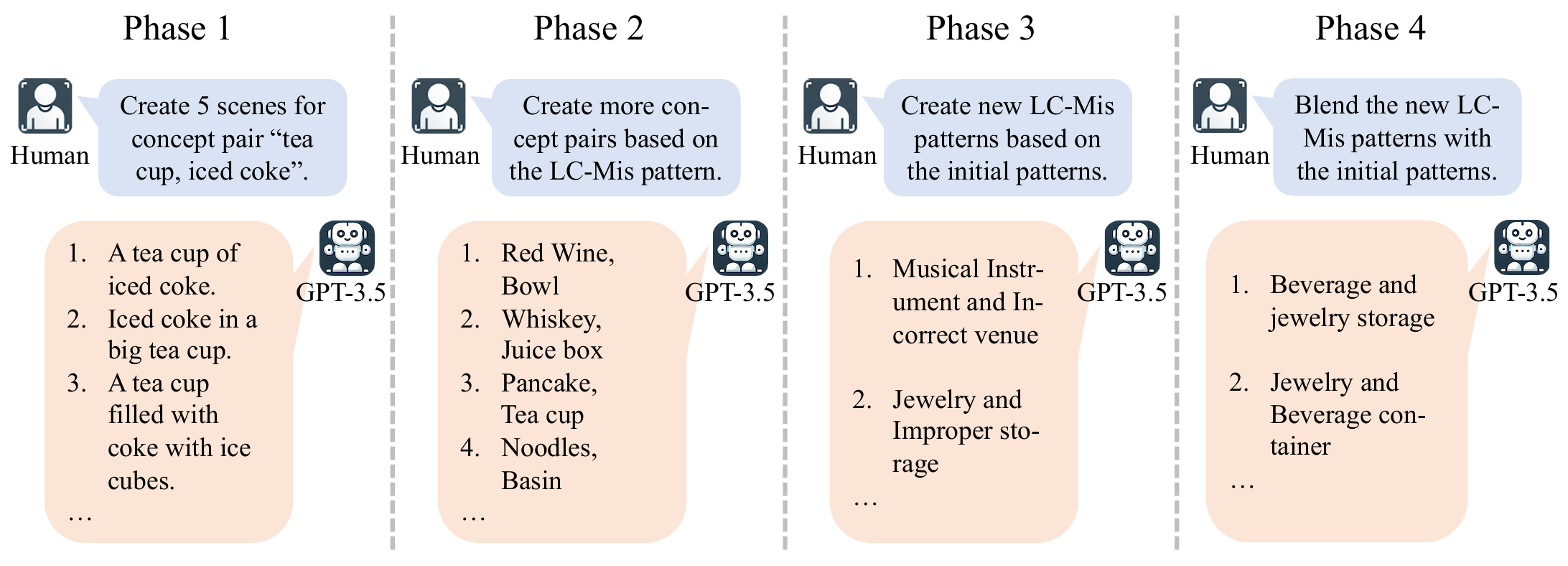}
    \caption{A single loop of our interactive LLMs (i.e., \gpt) guidance system, comprising 4 phases. Human researchers provide instruction for \gpt, then \gpt create new \ourProblem concept pairs and patterns for evaluation. We can iterate the phase 2 - 4 to rapidly expand our \ourProblem dataset.
    }
    \label{fig:DatasetGeneratingStructure}
\end{figure}

In this section, we detail the process of collecting our \ourProblem dataset. In our dataset, each pair features Concept \conceptA and Concept \conceptB, which cannot be simultaneously generated by text-to-image models due to the existance of the latent concept, \conceptC. The advantages of our collection system are as follows:

\begin{itemize}
    \item Given that extensive human knowledge is required to generate data with \ourProblem issue, we utilize LLMs (e.g., \gpt in this paper) to develop an efficient guidance system, inspired by LLMs Reasoning~\cite{dong2023large}.

    \

    \item These concept pairs often defy common human understanding, making them challenging to mine. We meticulously craft a generation loop containing 4 phases, allowing the iterative amplification of a small dataset into a substantially larger one. The process overview is shown in Figure~\ref{fig:DatasetGeneratingStructure}, and the detailed prompts to guide \gpt can be found in our Appendix Section~\ref{sec:sup_prompt_with_gpt}.

    \

    \item Finally, concept pairs discovered by \gpt require comprehensive evaluation. However, advanced tools such as Clipscore~\cite{hessel2021clipscore} or Image Reward~\cite{xu2023imagereward} may exhibit ineffectiveness in \ourProblem problem (Section~\ref{sec:analysis}). Therefore, the evaluation of human experts is an important part of our system.
\end{itemize}

\subsection{Phase 1 - Identifying Initial Concept Pairs as Seeds}
\label{phase1}

In Phase 1, human experts identify a small number of concept pairs and extract patterns from them. These patterns serve as seeds for subsequent dataset expansion.
Human researchers begin with analyzing various visual scenes in renowned text-to-image datasets, such as Laion2B-en\footnote{https://huggingface.co/datasets/laion/laion2B-en} and MJ User Prompts \& Images Dataset\footnote{https://huggingface.co/datasets/succinctly/midjourney-prompts}, to detect concept pairs prone to \ourDataset. They select 50 concept pairs, like \say{iced coke} and \say{tea cup}, which produce inaccurate output images.

Upon analysis, we categorize these concept pairs into 8 distinct patterns. Among them, 4 patterns belong to the scope of \ourProblem (detailed \ourDataset patterns are shown in Appendix Table~\ref{tab:category}), mainly including the categories of \say{foreground and background}, as well as \say{objects and containers}. These patterns serve as an effective starting point for identifying additional concept pairs by means of \gpt.

\subsection{Phase 2 - Generating and Verifying Additional Concept Pairs with \gpt}
\label{phase2}

In this phase, human researchers employ \gpt to generate additional concept pairs corresponding to each pattern from Section~\ref{phase1}, using the initial 50 concept pairs for few-shot learning. 
Using this method, \gpt generates 499 valid concept pairs. None of these pairs can be generated correctly by text-to-image models with the simple prompt \say{concept \conceptA, concept \conceptB} and they undergo rigorous verification, with most of them proving to be of high quality.

Then a further verification is  meticulously designed to comprehensively assess the accuracy of the generated concept pairs. Specifically, \gpt generates 5 prompts for each pair, varying in length and richness. Subsequently, text-to-image models generate 4 images per prompt. After human verification of all 20 images, the concept pairs are scored on 5 levels: zero correct images correspond to Level 5, $1\sim  5$ correct images to Level 4, $6\sim  10$ to Level 3, $11\sim  15$ to Level 2, and $16\sim  20$ to Level 1.

In total, 2,495 sentence-based text prompts are fed into text-to-image models, which in turn produce 9,980 images. After rigorous screening, 272 concept pairs (55\%) attain a Level 5 rating, representing the pinnacle of quality. Among them, 173 concept pairs belong to the scope of \ourProblem and will be used as the main data for our subsequent experiments. Utilizing \gpt's generalizing ability and reasoning skills, our initial dataset of 50 pairs is expanded quickly, maintaining high quality and diversity. This phase can be repeated on the basis of Phases 3 and 4 to generate more and concept pairs.

\subsection{Phase 3 - Discovering New Patterns for Concept Pair Generation}
\label{phase3}

In this phase, we leverage \gpt to identify 9 new patterns and apply the same methodology in Section~\ref{phase2} to these patterns.

During previous process, we observe that \gpt  produces duplicates when tasked with generating additional concept pairs, suggesting that it may have reached the limits of its knowledge within current patterns. Consequently, \gpt is encouraged to autonomously identify new patterns to enhance the semantic scope of \ourDataset. Building upon the 8 meticulously categorized patterns, human researchers instruct \gpt to identify 9 additional patterns. Subsequently, we replicate the procedures from Section~\ref{phase2}, achieving a Level 5 accuracy rate of 70\%, markedly exceeding the capabilities of human experts.

\subsection{Phase 4 - Creating Novel Concept Pairs by Merging Patterns}

In this phase, we merge concepts from different patterns to form new patterns and concept pairs, which will serve as seeds for the subsequent iteration.

Specifically, human researchers instruct \gpt to combine concepts from one pattern with ones from another pattern. For example, patterns labeled \say{Beverage and Incorrect Container} and \say{Jewelry and Inadequate Storage} are merged to create new patterns: \say{Beverage and Jewelry Storage} and \say{Jewelry and Beverage Container}. Following the process in Section~\ref{phase3}, we observe that each newly created pattern achieves a Level 5 accuracy rate of at least 60\%, demonstrating \gpt's capability in synthesizing orthogonal patterns into more patterns.

\

\noindent
\textbf{In summary}, we collect our dataset leveraging the collaboration between LLMs (i.e., \gpt) and text-to-image models. Based on our proposed system, we can further expand our dataset iteratively in the future.

\begin{figure}[t]
    \centering
    \includegraphics[width=\linewidth]{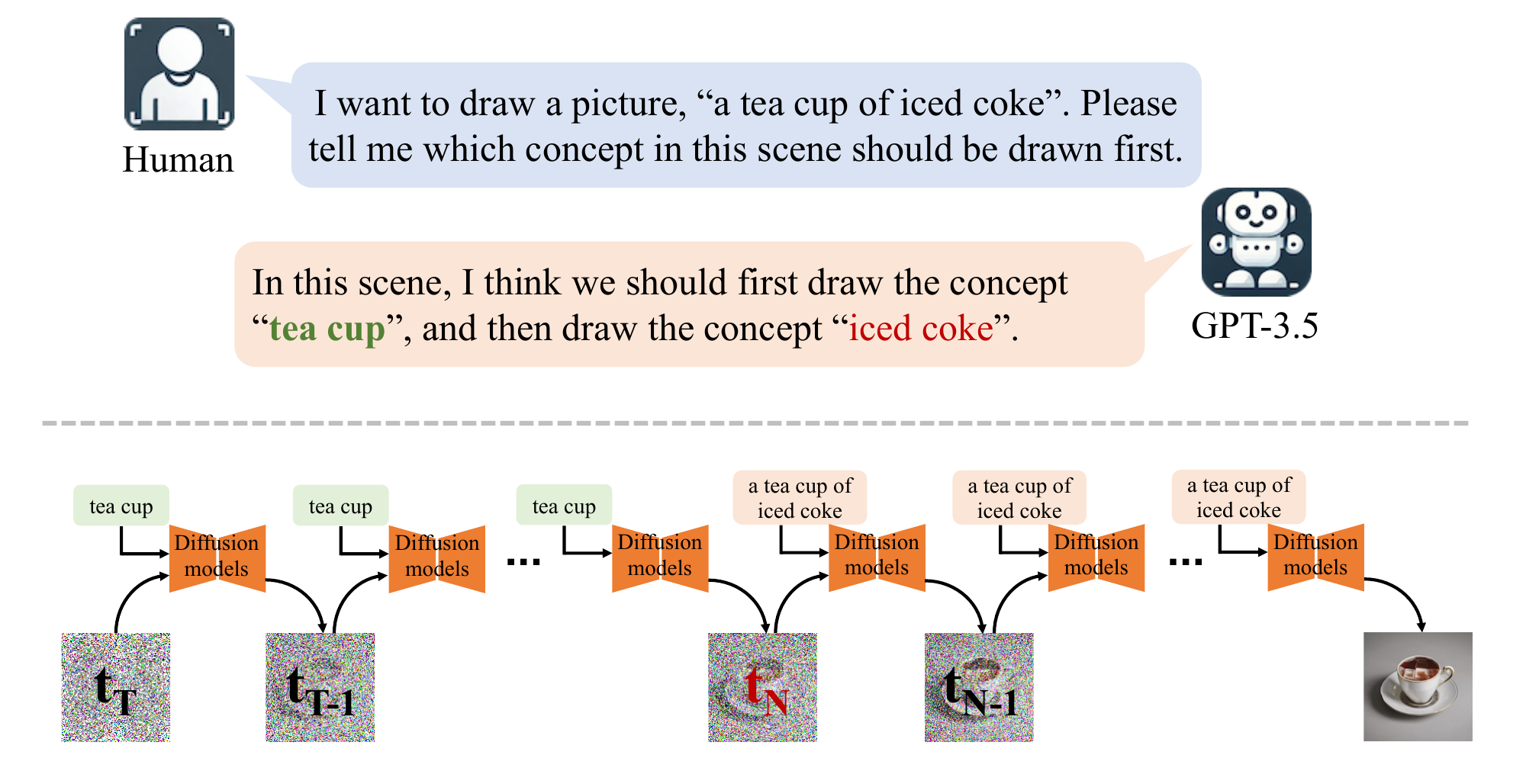}
    \caption{Overview of our method, \ourMethod. \gpt determines the drawing sequence of concepts. The initial concept is sampled for $N$ steps in diffusion models before inputting the full text prompt. We use binary search to find the optimal $N$, refer to Section~\ref{sec:our_method} Paragraph ``Dynamic Binary Optimization'' and Appendix Figure~\ref{fig:method1} and~\ref{fig:method2} for details.}
    \label{fig: method1}
\end{figure}

\section{Method: Mixture of Concept Experts (\dynamic)}
\label{sec:our_method}
\textbf{Motivation of \dynamic\ }
Humans always follow a certain order when painting. Inspired by human painting nature and motivated by dynamic models~\cite{han2021dynamic},
we integrate LLMs (e.g., \gpt in this paper) into our method, Mixture of Concept Experts (\dynamic) to alleviate the \ourProblem issues. 
As shown in Figure~\ref{fig: method1}, we first input the easily overlooked concept to focus on the attention mechanism during the early diffusion stages, enhancing their representation in the final image.
We base \dynamic on \sdxl~\cite{podell2023sdxl}, one of the foremost reliable open-source diffusion models. The system includes components listed as follows:
 \\
 \\
\noindent
\textbf{Sequential Concept Introduction\ }
Gaining insight from human artistic processes, concepts are introduced to diffusion models sequentially rather than simultaneously to prevent the \ourProblem entanglement. As a result, we need to find a reasonable and logical sequence of them.
LLMs are naturally suitable for this task.
Therefore, we employ \gpt to ascertain the most logical sequence for two concepts based on its comprehension of human behavior.
We provide interaction details with \gpt in our Appendix Section~\ref{sec:sup_prompt_with_gpt}.
 \\
 \\
\noindent
\textbf{Denoising Process Partition\ }
In the denoising process of diffusion models across time steps, $t_{T}, t_{T-1}, ..., t_1, t_0$, the process is split into 2 phases:
\begin{itemize}
    \item First Phase ($t_{T}, t_{T-1}, ..., t_{N}$): We only input the easily lost concept into diffusion models, and save images at each time step as a preparatory list.

     \ 
    
    \item Second Phase ($t_{N-1}, t_{N-2}, ..., t_{0}$): $t_x$ will be selected from the preparatory list. Starting from time step $t_{x}$, we provide the completed text prompt to diffusion models to synthesize the final image.
\end{itemize}
However, which image $t_x$ to take from the preparatory list reaches the optimal output remains unknown, motivating us to formulate a strategy for autonomously determining the optimal allocation of time steps.

\noindent
\textbf{Automated Feedback Mechanism\ }
We iteratively select the most suitable $t_x$ based on the fidelity of the final image. In this case, \textit{\clipscore}, $\mathcal{S}_c$, is used to assess the fidelity of the generated image $M$ to the desired concept, e.g., \conceptA:
\begin{equation}
    \mathcal{S}_c(M,\ \mathcal{A}) = ClipScore(M,\ \mathcal{A})
\end{equation}
As the issue of misalignment persists, naively employing \clipscore wouldn't truly show the correlation between entity $\mathcal{A}$ and the final image $M$ since it may be not distinguishable (see Section~\ref{sec:analysis}).
Therefore, we calculate the \clipscore between images and both the concept itself and its corresponding description generated by \gpt. 
Some details are not prominent in $M$, yet they might be further elaborated in the information further provided by the descriptions.
This possibly uplifts the \clipscore of the entity and lowers the influence of misalignment to some certain extent.
The the score function in \dynamic, \ourScore, is as follow:
\begin{equation}
    \mathcal{S}(M,\ \mathcal{A}) = Max(\mathcal{S}_c(M,\ \mathcal{A}),\ \mathcal{S}_c(M,\ \mathcal{A}_{Description}))
\end{equation}
We also devise a new metric named \ourMetric, denoted as $\mathcal{D}$:
\begin{equation}
    \mathcal{D} = \mathcal{S}(M, \mathcal{A}) - \mathcal{S}(M, \mathcal{B})
\end{equation}
$\mathcal{D}$ denotes the difference in the scores of an image between two concepts.
The larger the size of $|\mathcal{D}|$, the higher the likelihood that at least one concept within it will become blurred.

 \ 

\noindent
\textbf{Dynamic Binary Optimization\ }
When allocating a larger number of time steps in the first stage, the more dominant features of \conceptA possibly overshadows those of \conceptB, then the greater the absolute value of $\mathcal{D}$ is, and vice versa.
Therefore, there exists a positive correlation between the time steps allocated in the first phase and the image quality, making the binary search method highly suitable for this scenario. When the score of concept \conceptA is significantly higher than \conceptB, we choose a larger $t_x$ as the boundary point for two phases, and vice versa. If the difference of their scores is below the threshold, the final image will be output.
Specific implementation details can be found in our Appendix Section~\ref{sec:sup_moce}.
 \\
 \\
\noindent
\textbf{In summary}, our methodology presents a robust solution for addressing entity misalignment in text-to-image diffusion models. It combines intuitive reasoning, automatic fine-tuning, and efficiency optimizations to produce more precise and contextually apt image outputs.

\begin{figure*}[t]
    \begin{center}
        \adjustbox{max width=\linewidth}{
        \begin{tabular}{c}
            \begin{subfigure}{\linewidth}
                \captionsetup{font=normalsize}
                \centering
                \begin{tabular}{c | ccc}
                    \multicolumn{1}{c}{SDXL} & \multicolumn{3}{c}{MoCE} \\
                    \includegraphics[width=0.24\linewidth]{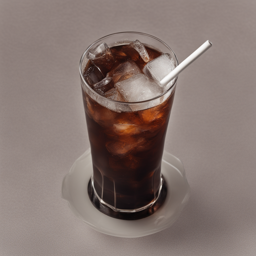} &
                    \includegraphics[width=0.24\linewidth]{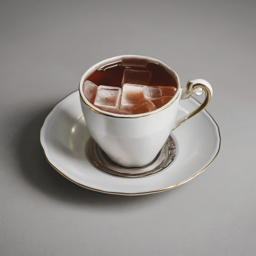} &
                    \includegraphics[width=0.24\linewidth]{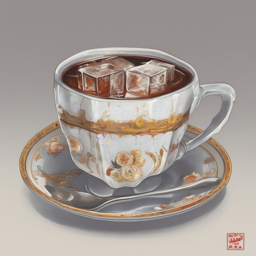} &
                    \includegraphics[width=0.24\linewidth]{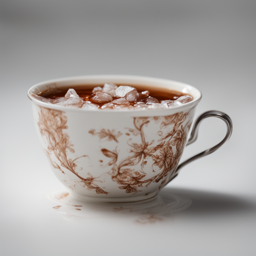} \\
                \end{tabular}
                \caption{Iced coke, tea cup}
            \end{subfigure} \\

            \begin{subfigure}{\linewidth}
                \captionsetup{font=normalsize}
                \centering
                \begin{tabular}{c | ccc}
                    \includegraphics[width=0.24\linewidth]{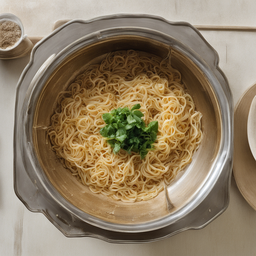} &
                    \includegraphics[width=0.24\linewidth]{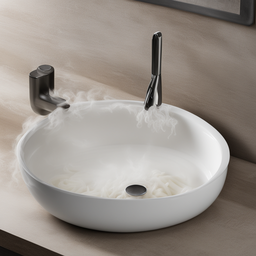} &
                    \includegraphics[width=0.24\linewidth]{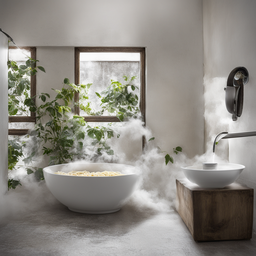} &
                    \includegraphics[width=0.24\linewidth]{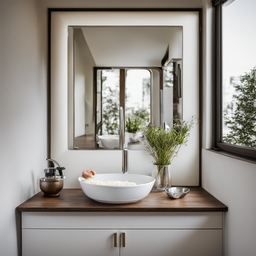} \\
                \end{tabular}
                \caption{Noodles, basin}
            \end{subfigure} \\

            \begin{subfigure}{\linewidth}
                \captionsetup{font=normalsize}
                \centering
                \begin{tabular}{c | ccc}
                    \includegraphics[width=0.24\linewidth]{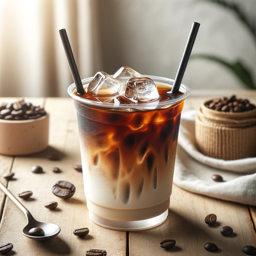} &
                    \includegraphics[width=0.24\linewidth]{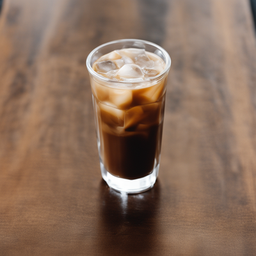} &
                    \includegraphics[width=0.24\linewidth]{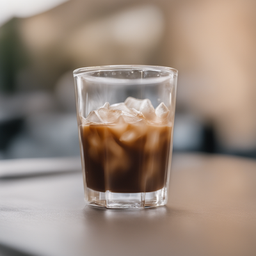} &
                    \includegraphics[width=0.24\linewidth]{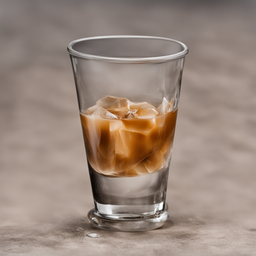} \\
                \end{tabular}
                \caption{Iced coffee, shot glass}
            \end{subfigure}
        \end{tabular}
        }
    \end{center}
    \caption{Visualizations of images at Level 5 generated by \sdxl (baseline) and \dynamic. Here, we present representative examples of the ``Item and Container'' pattern.}
    \label{fig:exp_visualization}
\end{figure*}

\section{Experiments}
\label{sec:exp}
We conduct extensive experiments centered around \dynamic, revealing its ability to alleviate the \ourProblem issue in text-to-image diffusion models.

\subsection{Setup}
\textbf{Dataset\ }
In Section~\ref{phase2}, we obtain 272 concept pairs for Level 5. From this pool, human experts carefully select 173 concept pairs as \ourProblem cases for experiments, while the remaining concept pairs fall into the realm of traditional misalignment issues, extensively discussed in Section~\ref{sec:related}.

\begin{figure*}[t]
    \begin{center}
        \adjustbox{max width=\linewidth}{
        \begin{tabular}{c}
            \begin{subfigure}{\linewidth}
                \captionsetup{font=normalsize}
                \centering
                \begin{tabular}{c | ccc}
                    \multicolumn{1}{c}{SDXL} & \multicolumn{3}{c}{MoCE} \\
                    \includegraphics[width=0.24\linewidth]{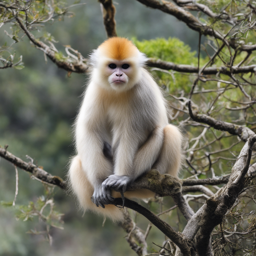} &
                    \includegraphics[width=0.24\linewidth]{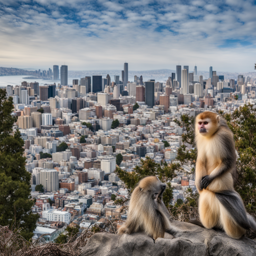} &
                    \includegraphics[width=0.24\linewidth]{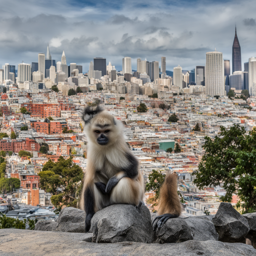} &
                    \includegraphics[width=0.24\linewidth]{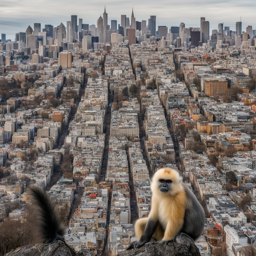} \\
                \end{tabular}
                \caption{Snubnosed monkey, San Fransisco}
            \end{subfigure} \\

            \begin{subfigure}{\linewidth}
                \captionsetup{font=normalsize}
                \centering
                \begin{tabular}{c | ccc}
                    \includegraphics[width=0.24\linewidth]{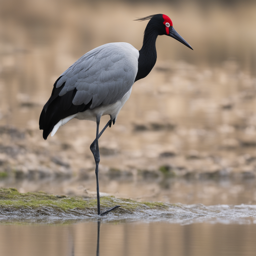} &
                    \includegraphics[width=0.24\linewidth]{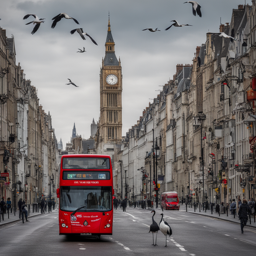} &
                    \includegraphics[width=0.24\linewidth]{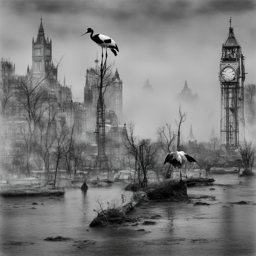} &
                    \includegraphics[width=0.24\linewidth]{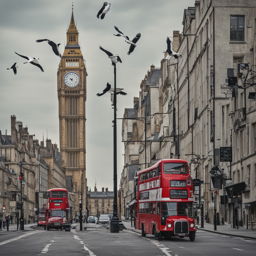} \\
                \end{tabular}
                \caption{Black necked crane, London}
            \end{subfigure} \\

            \begin{subfigure}{\linewidth}
                \captionsetup{font=normalsize}
                \centering
                \begin{tabular}{c | ccc}
                    \includegraphics[width=0.24\linewidth]{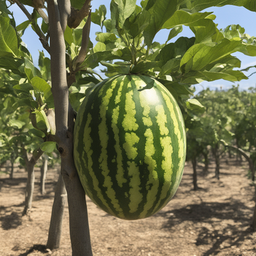} &
                    \includegraphics[width=0.24\linewidth]{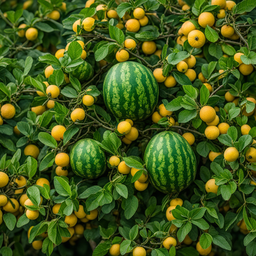} &
                    \includegraphics[width=0.24\linewidth]{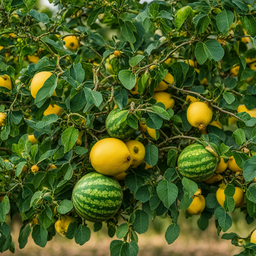} &
                    \includegraphics[width=0.24\linewidth]{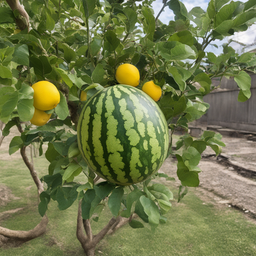} \\
                \end{tabular}
                \caption{Watermelon, lemon tree}
            \end{subfigure}
        \end{tabular}
        }
    \end{center}
    \caption{Visualizations of images at Level 5 generated by \sdxl (baseline) and \dynamic. Here, we present representative examples of the ``Foreground and Background'' pattern.}
    \label{fig:exp_visualization2}
\end{figure*}

\ 

\noindent
\textbf{Model\ }
Our remediation, \dynamic, is implemented using SDXL~\cite{podell2023sdxl} due to its open-source nature and widespread use. We omit \mj and Dall·E 3 from consideration because of their internal black-box architectures.
Our baseline, represented by these 173 \ourProblem concept pairs in Section~\ref{sec:dataset}, demonstrates that all of them receive a rating of Level 5, indicating that none of the images generated in 20 attempts faithfully represented the expected concepts.
We also incorporate 3 additional baseline models: Attend-and-Excite (AAE)\cite{chefer2023attend}, Dall·E 3\cite{openai2023dalle3}, and Anole~\cite{anole}. AAE aims to mitigate traditional misalignment issues via the attention map layer, and Dall·E 3 through fine-grained annotation. And Anole is a novel model that employs autoregressive methods for interleaved image-text generation, offering fresh insights into resolving misalignment issues.

Our experiments use a single NVIDIA A100 GPU for image generation via SDXL, and a RTX 4090 GPU is also sufficient.

 \

\noindent
\textbf{Evaluation Metric\ }
Considering the instability of quantitative score evaluation, as well as the use of \clipscore and \imgreward in our method, \dynamic,we primarily utilize human evaluation in our experiments. We engage human experts for a more impartial evaluation. Additionally, the results of quantitative score evaluation are also included in our Appendix Section~\ref{sec:sup_score_eval} for reference.

\begin{figure}[t]
    \centering
    \includegraphics[width=\linewidth]{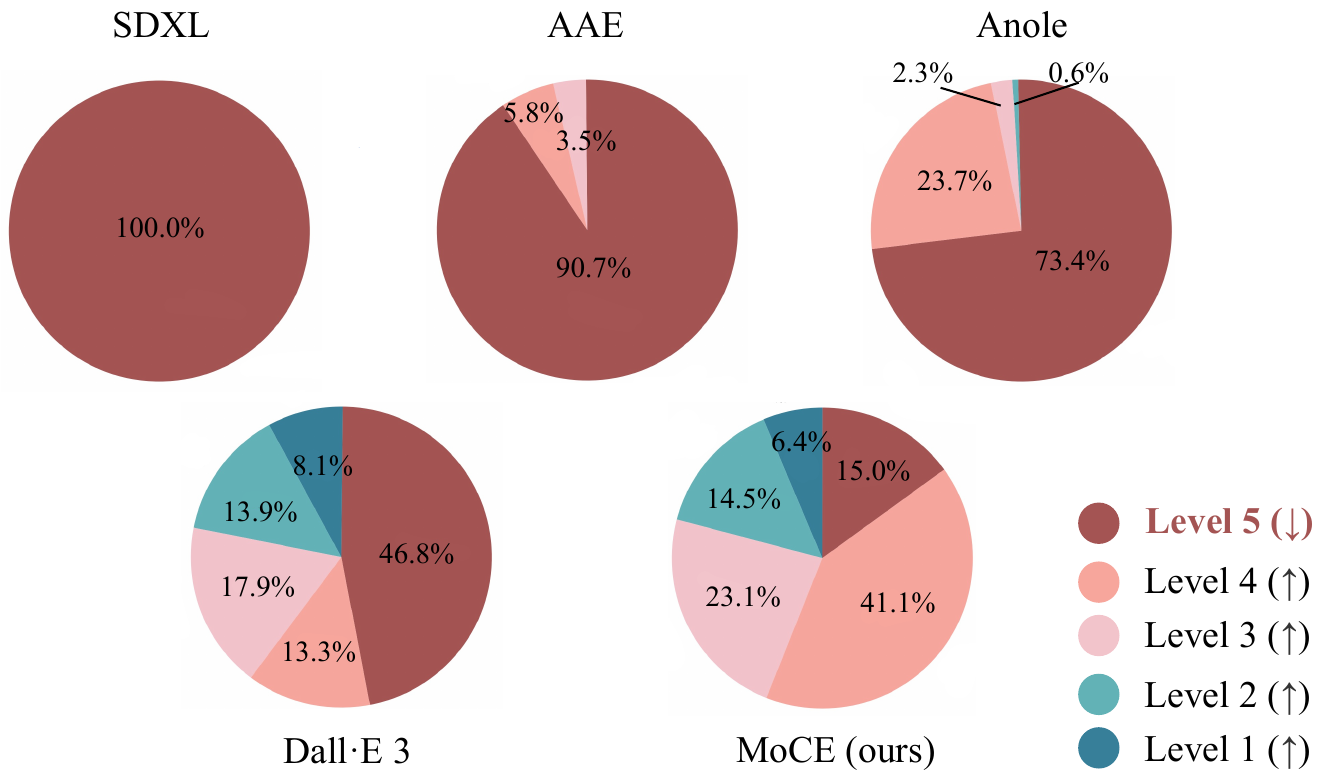}
    \caption{Human Evaluation for our \dynamic. Human experts rate the 173 \ourProblem concept pairs on \sdxl, AAE, Anole, Dalle·E 3 and our \dynamic. The \textcolor{levelfivebrown}{\textbf{brown}} color represents the proportion of Level 5 ratings among 173 pairs. A lower Level 5 proportion indicates the model better alleviates the \ourProblem issue. Compared to baseline models, our \dynamic exhibits a clear advantage when faced with \ourProblem issues, and it has even surpassed Dalle·E 3, which requires expensive training costs.}
    \label{fig:exp_base_result}
\end{figure}

\subsection{Result}
\label{exp:results}
Our method, \dynamic, utilizes the 173 Level 5 \ourProblem concept pairs, providing their corresponding text prompts as input to the model. The second phase of \dynamic is repeated up to three times to ensure the \ourMetric $\mathcal{D}$ coverage below 0.6. After obtaining 20 images for each concept pair, human experts re-evaluate these images based on the criteria discussed in Section~\ref{sec:dataset}. We report the counts of concept pairs at each level after undergoing improvement by our \dynamic and compare them to the baseline results in Figure~\ref{fig:exp_base_result}.
Meanwhile, we also present several visualized images in Figure~\ref{fig:exp_visualization} and~\ref{fig:exp_visualization2}.

\begin{figure}[t]
    \centering
    \includegraphics[width=\linewidth]{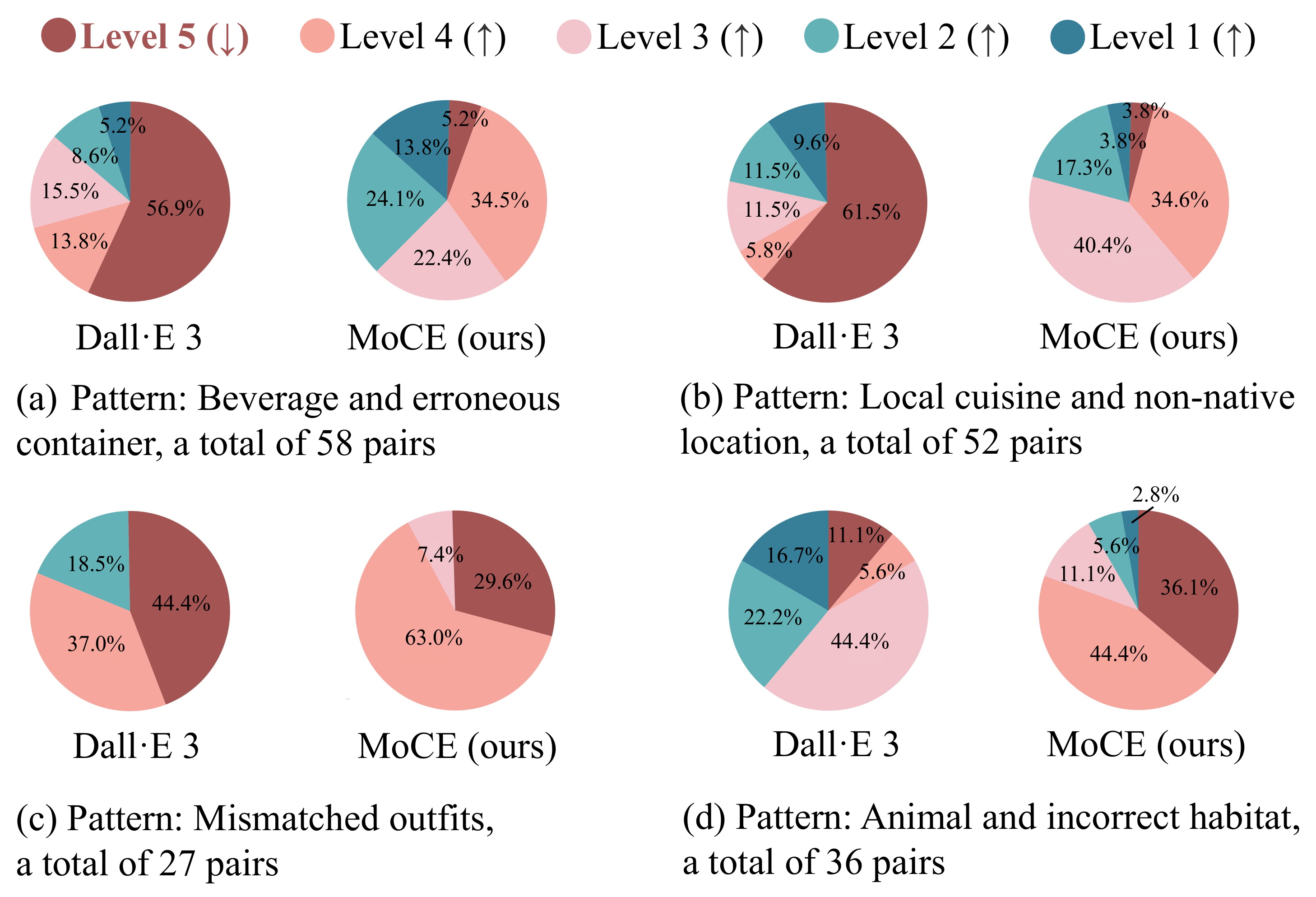}
    \caption{Human Evaluation for Dall·E 3 (baseline) and our \dynamic on different patterns mentioned in Section~\ref{sec:dataset}. The \textcolor{levelfivebrown}{\textbf{brown}} color represents the proportion of Level 5 ratings among 173 pairs. A lower Level 5 proportion indicates the model better alleviates the \ourProblem issue. In Patterns (a), (b), and (c), \dynamic outperforms Dall·E 3, as its output includes more concept pairs belonging to Level 1 and 2. And in Pattern (d), Dall·E 3's performance is superior to \dynamic.}
    \label{fig:pattern_result}
\end{figure}

In the original Level 5 \ourProblem concept pairs, the baseline model fails to produce any correct image, as we mentioned in Section~\ref{sec:dataset}. Even when sophisticated engineering strategies, e.g., AAE and Anole, are used, which may be effective for traditional misalignment problems, there is little improvement for Level 5 concept pairs when applied to \ourProblem problems. However, following the enhancement made by our \dynamic, over half of the concept pairs are now correctly generated, and there are even several concept pairs rated as Level 1. Additionally, Dall·E 3, with its expensive and fine-tuned data annotations, indeed helps with the \ourProblem problem, achieving improvements for Level 5 concept pairs comparable to \dynamic. However, it is important to note that training Dall·E 3 requires additional data preprocessing, which can be a costly process, suggesting that our \dynamic produces more correct images in an economical manner. %
We also include a comparison of the scores of images generated by \dynamic and baselines in Appendix Table~\ref{tab:score_eval}, demonstrating that our method is statistically meaningful.

In addition, we also present the human evaluation results under the segmented patterns. As introduced in Section~\ref{sec:dataset}, we have divided these 173 \ourProblem concept pairs into 4 patterns, and we have presented the results of human evaluation in Figure~\ref{fig:pattern_result}. In Patterns (a), (b) and (c), \dynamic outperforms Dall·E 3, as its output includes more concept pairs belonging to Level 1 and 2. And in Pattern (d), Dall·E 3's performance is superior to \dynamic.

\subsection{Analysis}
\label{sec:analysis}

\begin{figure}[t]
    \centering
    \includegraphics[width=\linewidth]{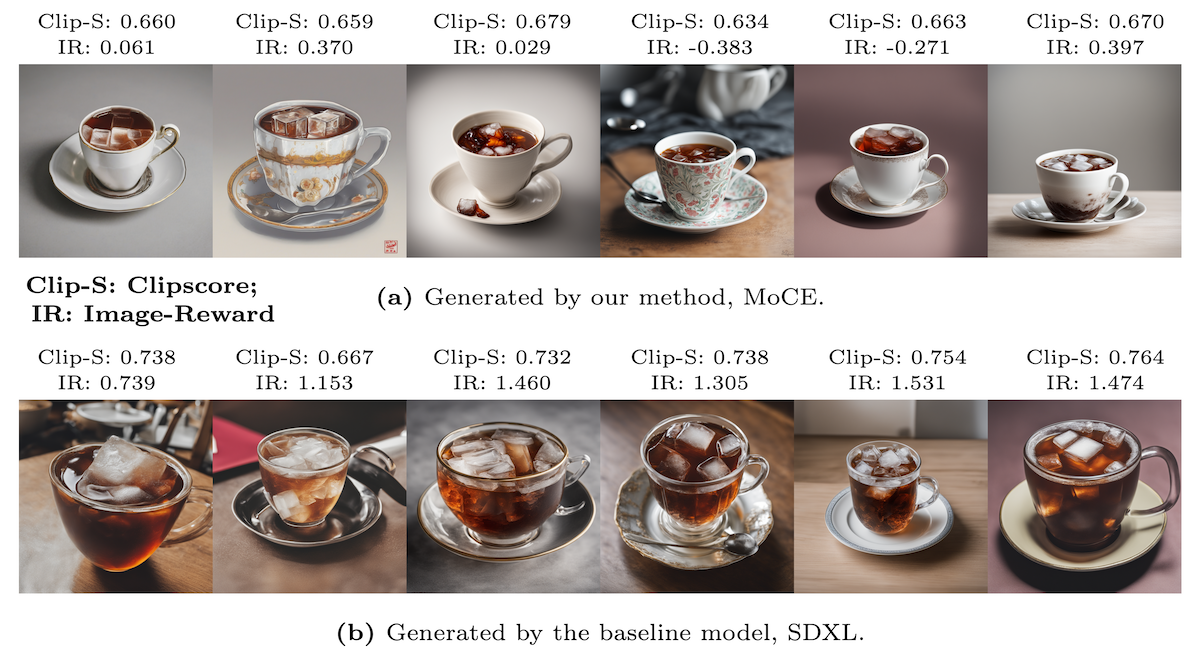}
  \caption{
      Visualization of images of ``a tea cup of iced coke'' generated by both our \dynamic and baseline model. We also report \clipscore (Clip-S, $\uparrow$) and \imgreward Score (IR, $\uparrow$) between images and the concept, ``iced coke'', to demonstrate the minor pitfalls of the existing evaluation metrics: even \dynamic correctly generates ``iced coke'' (a), the result score is still significantly lower.
    }
  \label{fig:check_benchmark}
\end{figure}

We are deeply concerned about the impact of \ourProblem issues on the existing text-to-image model landscape, for the most representative and challenging example, \say{a tea cup of iced coke}, we present visual restoration results in Figure~\ref{fig:check_benchmark}. Both \clipscore and \imgreward effectively adjust the time step in our \dynamic model with the help of \ourScore, while unfortunately the existing evaluation metrics may sometimes be proven to be not absolutely accurate enough. Figure~\ref{fig:check_benchmark} displays both \clipscore and \imgreward scores between images and the concept \say{iced coke}. We meticulously choose transparent glasses generated by baseline models, which resemble tea cups, to analyze the error in the scoring mechanism. The occurrence of \say{iced coke} in both sub-figures within Figure~\ref{fig:check_benchmark} is evident to human experts. However, both \clipscore and \imgreward scores are markedly lower for images of \say{a tea cup of iced coke} attributed to the material of the cup. This highlights the limitations of current evaluation metrics in addressing misalignment issues and emphasizes the necessity for developing new metrics based on existing methods, which will become one of the key research directions for our future work.

\begin{figure}[t]
    \centering
    \begin{subfigure}[b]{0.45\linewidth}
        \includegraphics[width=\textwidth]{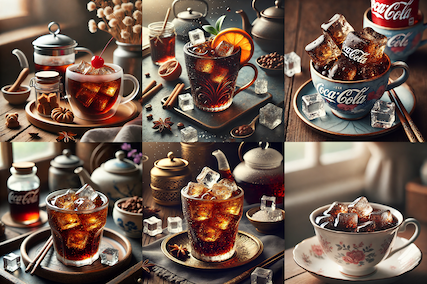}
        \caption{Dall·E 3 without GPT-4}
        \label{fig:new_dalle}
    \end{subfigure}
    \hfill
    \begin{subfigure}[b]{0.45\linewidth}
        \includegraphics[width=\textwidth]{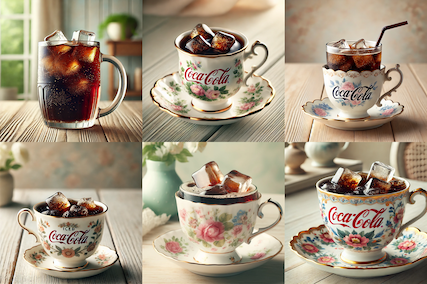}
        \caption{Dall·E 3 with GPT-4}
        \label{fig:new_dalle_chat}
    \end{subfigure}

    \begin{subfigure}[b]{0.45\linewidth}
        \includegraphics[width=\textwidth]{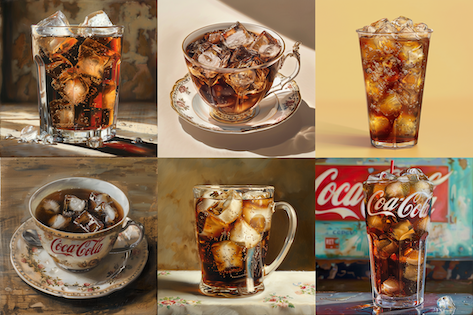}
        \caption{Midjourney}
        \label{fig:new_mid}
    \end{subfigure}
    \hfill
    \begin{subfigure}[b]{0.45\linewidth}
        \includegraphics[width=\textwidth]{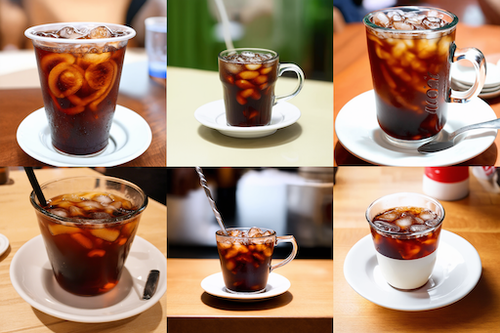}
        \caption{Stable Diffusion 3}
        \label{fig:new_sd3}
    \end{subfigure}

    \caption{The results of the example ``a tea cup of iced coke'' using the latest Dall·E 3 (a), Dall·E 3 with complex prompt engineering from GPT-4 (b), Midjourney(c) and Stable Diffusion 3 (d). The latest Dall·E 3 with GPT-4 has indeed alleviated the \ourProblem issue to some degree, with the help of significant annotation costs.}
    \label{fig:new_model}
\end{figure}

\section{Conclusions}
\subsection{Summary}
In this paper, we introduce a new text-to-image misalignment issue called \ourProblem, involving a significant latent concept. We present a novel framework to explore the \ourProblem examples, making it one of the earliest works of using LLMs in developing image generation systems. Our method, \dynamic, innovatively splits the text prompts and inputs them into diffusion models, alleviating the \ourProblem problem. Human evaluation confirms \dynamic's effectiveness. we also highlight the impact of the \ourProblem on existing text-to-image systems, discuss flaws in current evaluation metrics, and call for community attention to this matter.

\subsection{Recent Advances in the Field}
\label{sec:new_model}
Until releasing our paper, text-to-image models have further evolved. We present results from the latest (available online as of July 7, 2024) Dall·E 3, Midjourney, and Stable Diffusion 3 in Figure~\ref{fig:new_model}. In the example \say{a tea cup of iced coke},
Without complex prompt engineering, models still perform poorly on the \ourProblem issue, as shown in Figure~\ref{fig:new_dalle},~\ref{fig:new_mid} and~\ref{fig:new_sd3}. Complex prompt engineering from GPT-4 (Figure~\ref{fig:new_dalle_chat}) does help alleviate the issue. However, it's important to note that this comes with significant annotation costs during Dall·E 3's training, and is also accompanied by a certain degree of instability, highlighting the issue's significance.

\subsection{Future Work}
In our future work, we will focus on exploring more complex \ourProblem scenarios and developing learnable search algorithms to reduce the iterations in our method. Additionally, we will expand the range of model types, model versions and sampler types used in our dataset and continuously iterate our dataset collection algorithms to enhance and enlarge the dataset.

\section*{Acknowledgement}
This work was supported by Key R\&D Program of Shandong Province, China (2023CXGC010112).

\bibliographystyle{splncs04}
\bibliography{main}

\begin{thebibliography}{10}
\providecommand{\url}[1]{\texttt{#1}}
\providecommand{\urlprefix}{URL }
\providecommand{\doi}[1]{https://doi.org/#1}

\bibitem{achiam2023gpt}
Achiam, J., Adler, S., Agarwal, S., Ahmad, L., Akkaya, I., Aleman, F.L., Almeida, D., Altenschmidt, J., Altman, S., Anadkat, S., et~al.: Gpt-4 technical report. arXiv preprint arXiv:2303.08774  (2023)

\bibitem{brown2020language}
Brown, T., Mann, B., Ryder, N., Subbiah, M., Kaplan, J.D., Dhariwal, P., Neelakantan, A., Shyam, P., Sastry, G., Askell, A., et~al.: Language models are few-shot learners. In: NeurIPS (2020)

\bibitem{chefer2023attend}
Chefer, H., Alaluf, Y., Vinker, Y., Wolf, L., Cohen-Or, D.: Attend-and-excite: Attention-based semantic guidance for text-to-image diffusion models. ACM Transactions on Graphics (TOG)  \textbf{42}(4),  1--10 (2023)

\bibitem{anole}
Chern, E., Su, J., Ma, Y., Liu, P.: Anole: An open, autoregressive and native multimodal models for interleaved image-text generation. GitHub repository  (2024), \url{https://github.com/GAIR-NLP/anole}

\bibitem{couairon2022diffedit}
Couairon, G., Verbeek, J., Schwenk, H., Cord, M.: Diffedit: Diffusion-based semantic image editing with mask guidance. arXiv preprint arXiv:2210.11427  (2022)

\bibitem{dhariwal2021diffusion}
Dhariwal, P., Nichol, A.: Diffusion models beat gans on image synthesis. In: NeurIPS (2021)

\bibitem{ding2021cogview}
Ding, M., Yang, Z., Hong, W., Zheng, W., Zhou, C., Yin, D., Lin, J., Zou, X., Shao, Z., Yang, H., et~al.: Cogview: Mastering text-to-image generation via transformers. In: NeurIPS (2021)

\bibitem{dong2023large}
Dong, Q., Dong, L., Xu, K., Zhou, G., Hao, Y., Sui, Z., Wei, F.: Large language model for science: A study on p vs. np. arXiv preprint arXiv:2309.05689  (2023)

\bibitem{du2023reduce}
Du, Y., Durkan, C., Strudel, R., Tenenbaum, J.B., Dieleman, S., Fergus, R., Sohl-Dickstein, J., Doucet, A., Grathwohl, W.S.: Reduce, reuse, recycle: Compositional generation with energy-based diffusion models and mcmc. In: ICML (2023)

\bibitem{han2021dynamic}
Han, Y., Huang, G., Song, S., Yang, L., Wang, H., Wang, Y.: Dynamic neural networks: A survey. TPAMI  (2021)

\bibitem{hessel2021clipscore}
Hessel, J., Holtzman, A., Forbes, M., Bras, R.L., Choi, Y.: Clipscore: A reference-free evaluation metric for image captioning. arXiv preprint arXiv:2104.08718  (2021)

\bibitem{ho2022imagen}
Ho, J., Chan, W., Saharia, C., Whang, J., Gao, R., Gritsenko, A., Kingma, D.P., Poole, B., Norouzi, M., Fleet, D.J., et~al.: Imagen video: High definition video generation with diffusion models. arXiv preprint arXiv:2210.02303  (2022)

\bibitem{ho2020denoising}
Ho, J., Jain, A., Abbeel, P.: Denoising diffusion probabilistic models. In: NeurIPS (2020)

\bibitem{kawar2023imagic}
Kawar, B., Zada, S., Lang, O., Tov, O., Chang, H., Dekel, T., Mosseri, I., Irani, M.: Imagic: Text-based real image editing with diffusion models. In: CVPR (2023)

\bibitem{li2019controllable}
Li, B., Qi, X., Lukasiewicz, T., Torr, P.: Controllable text-to-image generation. In: NeurIPS (2019)

\bibitem{li2023gligen}
Li, Y., Liu, H., Wu, Q., Mu, F., Yang, J., Gao, J., Li, C., Lee, Y.J.: Gligen: Open-set grounded text-to-image generation. In: CVPR (2023)

\bibitem{liu2022compositional}
Liu, N., Li, S., Du, Y., Torralba, A., Tenenbaum, J.B.: Compositional visual generation with composable diffusion models. In: ECCV (2022)

\bibitem{mansimov2015generating}
Mansimov, E., Parisotto, E., Ba, J.L., Salakhutdinov, R.: Generating images from captions with attention. arXiv preprint arXiv:1511.02793  (2015)

\bibitem{meng2021sdedit}
Meng, C., He, Y., Song, Y., Song, J., Wu, J., Zhu, J.Y., Ermon, S.: Sdedit: Guided image synthesis and editing with stochastic differential equations. arXiv preprint arXiv:2108.01073  (2021)

\bibitem{midjourney}
{Midjourney}: {Midjourney (V5.2) [Text-to-Image Model]}. \url{https://www.midjourney.com} (2023)

\bibitem{nichol2021glide}
Nichol, A., Dhariwal, P., Ramesh, A., Shyam, P., Mishkin, P., McGrew, B., Sutskever, I., Chen, M.: Glide: Towards photorealistic image generation and editing with text-guided diffusion models. arXiv preprint arXiv:2112.10741  (2021)

\bibitem{chatgpt}
{OpenAI}: {ChatGPT (Aug 3 Version) [Large Language Model]}. \url{https://chat.openai.com} (2023)

\bibitem{openai2023dalle3}
OpenAI: Dall·e 3 system card. OpenAI technical report  (2023)

\bibitem{podell2023sdxl}
Podell, D., English, Z., Lacey, K., Blattmann, A., Dockhorn, T., M{\"u}ller, J., Penna, J., Rombach, R.: Sdxl: Improving latent diffusion models for high-resolution image synthesis. arXiv preprint arXiv:2307.01952  (2023)

\bibitem{ramesh2022hierarchical}
Ramesh, A., Dhariwal, P., Nichol, A., Chu, C., Chen, M.: Hierarchical text-conditional image generation with clip latents. arXiv preprint arXiv:2204.06125  (2022)

\bibitem{ramesh2021zero}
Ramesh, A., Pavlov, M., Goh, G., Gray, S., Voss, C., Radford, A., Chen, M., Sutskever, I.: Zero-shot text-to-image generation. In: ICML (2021)

\bibitem{reed2016generative}
Reed, S., Akata, Z., Yan, X., Logeswaran, L., Schiele, B., Lee, H.: Generative adversarial text to image synthesis. In: ICML (2016)

\bibitem{rombach2022high}
Rombach, R., Blattmann, A., Lorenz, D., Esser, P., Ommer, B.: High-resolution image synthesis with latent diffusion models. In: CVPR (2022)

\bibitem{ronneberger2015u}
Ronneberger, O., Fischer, P., Brox, T.: U-net: Convolutional networks for biomedical image segmentation. In: MICCAI (2015)

\bibitem{saharia2022photorealistic}
Saharia, C., Chan, W., Saxena, S., Li, L., Whang, J., Denton, E.L., Ghasemipour, K., Gontijo~Lopes, R., Karagol~Ayan, B., Salimans, T., et~al.: Photorealistic text-to-image diffusion models with deep language understanding. In: NeurIPS (2022)

\bibitem{schuhmann2022laion}
Schuhmann, C., Beaumont, R., Vencu, R., Gordon, C., Wightman, R., Cherti, M., Coombes, T., Katta, A., Mullis, C., Wortsman, M., et~al.: Laion-5b: An open large-scale dataset for training next generation image-text models. In: NeurIPS (2022)

\bibitem{sharma2018conceptual}
Sharma, P., Ding, N., Goodman, S., Soricut, R.: Conceptual captions: A cleaned, hypernymed, image alt-text dataset for automatic image captioning. In: ACL (2018)

\bibitem{song2023consistency}
Song, Y., Dhariwal, P., Chen, M., Sutskever, I.: Consistency models. arXiv preprint arXiv:2303.01469  (2023)

\bibitem{vaswani2017attention}
Vaswani, A., Shazeer, N., Parmar, N., Uszkoreit, J., Jones, L., Gomez, A.N., Kaiser, {\L}., Polosukhin, I.: Attention is all you need. In: NeurIPS (2017)

\bibitem{wang2023compositional}
Wang, R., Chen, Z., Chen, C., Ma, J., Lu, H., Lin, X.: Compositional text-to-image synthesis with attention map control of diffusion models. arXiv preprint arXiv:2305.13921  (2023)

\bibitem{wu2022nuwa}
Wu, C., Liang, J., Hu, X., Gan, Z., Wang, J., Wang, L., Liu, Z., Fang, Y., Duan, N.: Nuwa-infinity: Autoregressive over autoregressive generation for infinite visual synthesis. arXiv preprint arXiv:2207.09814  (2022)

\bibitem{xu2023imagereward}
Xu, J., Liu, X., Wu, Y., Tong, Y., Li, Q., Ding, M., Tang, J., Dong, Y.: Imagereward: Learning and evaluating human preferences for text-to-image generation. arXiv preprint arXiv:2304.05977  (2023)

\bibitem{xu2017attngan}
Xu, T., Zhang, P., Huang, Q., Zhang, H., Gan, Z., Huang, X., He, X.: Attngan: Fine-grained text to image generation with attentional generative adversarial networks. arxiv. arXiv preprint arXiv:1711.10485  (2017)

\bibitem{yu2022scaling}
Yu, J., Xu, Y., Koh, J.Y., Luong, T., Baid, G., Wang, Z., Vasudevan, V., Ku, A., Yang, Y., Ayan, B.K., et~al.: Scaling autoregressive models for content-rich text-to-image generation. arXiv preprint arXiv:2206.10789  (2022)

\bibitem{zhang2017stackgan}
Zhang, H., Xu, T., Li, H., Zhang, S., Wang, X., Huang, X., Metaxas, D.N.: Stackgan: Text to photo-realistic image synthesis with stacked generative adversarial networks. In: ICCV (2017)

\end{thebibliography}

\newpage

\appendix
\section*{Appendix}
\section{Interaction Details in Interactive LLMs Guidance System}
\label{sec:sup_prompt_with_gpt}

\begin{figure*}[h]
    \begin{center}
        \adjustbox{max width=\linewidth}{
            \begin{tabular}{c c c c}
            \includegraphics[height=5cm]{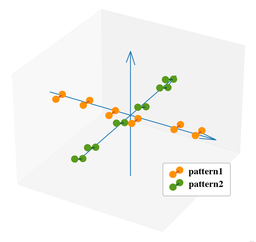} &
            \includegraphics[height=5cm]{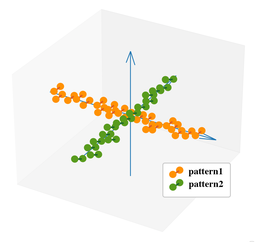} &
            \includegraphics[height=5cm]{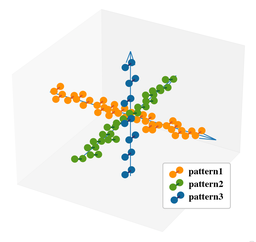} &
            \includegraphics[height=5cm]{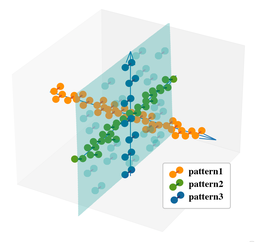} \\
            (a) Round 1  & (b) Round 2-3 & (c) Round 4-5 & (d) Round 6 \\
            Come up with concept pairs& Guide \gpt to extend concept pairs & Guide \gpt to create new patterns& Guide \gpt to blend exsiting patterns \\
            \end{tabular}
        }
    \end{center}
    \caption{Schematic diagrams of different rounds of \socratic. The complete procedure contains coming up with concept pairs (Figure a), guiding \gpt to extend concept pairs (Figure b), guiding \gpt to create new patterns (Figure c) and guiding \gpt to blend exsiting patterns (Figure d).
    }
    \label{fig:socrate}
\end{figure*}

In this section, we describe the detailed process of interaction between human researchers and \gpt within our workflow. Utilizing block diagrams for clarity, we model the engagement between human experts and \gpt.
To illustrate the variation in data volume within the Interactive LLMs Guidance System, we present several diagrams, in Figure~\ref{fig:socrate}. In these diagrams, pairs of spheres along the same axis denote concept pairs associated with a particular pattern, distinguishable by their color.

\textbf{Phase 1: Identifying Initial Concept Pairs as Seeds}
After developing an initial set of 50 concept pairs and categorizing them into 8 unique patterns (among them, 4 patterns match our \ourProblem issue), human researchers direct \gpt to generate additional concept pairs. We specify both the researcher-provided prompt and the \gpt response. 
The format of the patterns deducting concept pairs are depicted in Table \ref{tab:category}.

Midjourney text-to-image generation, in conjunction with human researcher validation, then confirms a total of 50 concept pairs.
Throughout the evaluation conducted by human experts, 5 impartial human experts are tasked to identify the generated images. If minor differences, a majority opinion will be adopted. Otherwise, a senior expert will come to re-identify the image. Human expert evaluation prove to be accurate under certain circumstances as discussed in Section~\ref{sec:analysis} of our paper.

\begin{table}[ht]
    \caption{Patterns summarized manually by human researchers.}
    \label{tab:category}
    \centering
    \begin{tabular}{@{}cc@{}}
        \toprule
        \multicolumn{2}{c}{Pattern}                                          \\ \midrule
        Beverage and erroneous container      & Mismatched outfits           \\
        Local cuisine and non-native location & Animal and incorrect habitat \\
        \multicolumn{2}{c}{...}                                              \\ \bottomrule
\end{tabular}
\end{table}

\begin{tcolorbox}[colback=blue!5!white,colframe=blue!75!black]
\small
{\bf Prompt:}  I want you to act as an AI creative assistant helping users generate pairs of nouns on cue: pairs of Beverage and erroneous container (e.g. iced coke, tea cup). Each pair occupies one line, in 1 \~{} 3 words, separated by commas.
\\
{\bf Answer:}
Certainly, I can help you generate pairs of nouns as requested. Here are some pairs of beverages and wrong containers:
\vspace{2mm}
\\1. Coffee, Champagne Flute
\\2. Orange Juice, Coffee Mug
\\3. Hot Chocolate, Martini Glass
\\4. Iced Tea, Shot Glass
\\5. Lemonade, Beer Stein
\\......
\vspace{2mm}
\\These pairs should provide some creative and humorous combinations of beverages and the erroneous containers for your use. If you need more pairs or have any specific preferences, feel free to let me know!
\end{tcolorbox}

We leverage \gpt to generate 5 detailed prompts for each concept pair:

\begin{tcolorbox}[colback=blue!5!white,colframe=blue!75!black]
\small {\bf Prompt:} 
I want you to act as an AI creative assistant helping users generate images using Midjourney. Your job is to generate a prompt for text-to-image generation model. You are good at text-image concept grounding, visual concept understanding, text-to-image diffusion models' prompt engineering. Don't be verbose in your answers, only provide the proposed prompt itself and avoid including explanations. I've provided you with a pair of visual concepts, and I want you to provide me with detailed prompts to feed into Midjourney that one given concept would show up with the other one in the corresponding generated image. It will be your job to come up with five prompts that closely integrate two concepts visually, yet each remains a distinct entity. Remember there is NO extra object or human in the prompt, NOR any explanatory sentences, according to the user's specifications.
\\Now, the given two visual concepts are: Iced coke, tea cup
\\
{\bf Answer:}
\\1. Ice coke bubbling in a tea cup.
\\2. A tea cup of ice coke.
\\3. Ice coke poured into a tea cup.
\\4. Iced coke in a tea cup.
\\5. Tea cups filled with iced coke, cold fogging up.
\end{tcolorbox}

\newpage

\textbf{Phase 2 - Generating and Verifying Additional Concept Pairs}
On the basis of known patterns, we encourage \gpt to explore more concept pairs. Our instructions for \gpt are as follows:

\begin{tcolorbox}[colback=blue!5!white,colframe=blue!75!black]
\small {\bf Prompt:} I want you to act as a concept pair generator for text-to-image generation model, midjourney. You are good at text-image concept grounding, visual concept understanding, text-to-image diffusion models' prompt engineering. Don't be verbose in your answers, but do provide details and examples where it might help the explanation. 
I’ve provided you with a bunch of existing positive and negative noun-concept pairs. The existing positive pairs comprise a common beverage and an unconventional common container that is not typically used to hold or serve that beverage, making them positive pairs. Specifically, given noun-concept pair like "iced coke" and "tea cup", they are both common in the real world. However, the resultant text-to-image synthesis unexpectedly refers to "a glass cup containing iced coke", with the concept of "tea cup" ignored. It will be your job to come up with 30 such new positive noun-concept pairs  following the given pairs. You need to remember that nouns cannot be repeated between different concept pairs.

\scriptsize

\vspace{1mm}
\begin{minipage}{0.45\textwidth}
    \PositiveFewShot Coffee, Champagne Flute

    \PositiveFewShot Orange Juice, Coffee Mug
    
    \PositiveFewShot Iced Tea, Shot Glass
    
    \PositiveFewShot Milkshake, Whiskey Tumbler
    
    \PositiveFewShot Coca-Cola, Water Bottle
    
    \PositiveFewShot Green Tea, Red Solo Cup
    
    \PositiveFewShot Hot Coffee, Cocktail Shaker
    
    \PositiveFewShot Iced Coffee, Shot Glass
    
    \PositiveFewShot Pina Colada, Beer Stein
    
    \PositiveFewShot Iced Latte, Soda Can
    
    \PositiveFewShot Espresso, Water Bottle
    
    \PositiveFewShot Hot Tea, Coffee Mug
    
    \PositiveFewShot Apple Cider, Cocktail Shaker
    
    \PositiveFewShot Hot Chocolate, Red Solo Cup
    
    \PositiveFewShot Pumpkin Latte, Water Bottle
    
    \PositiveFewShot Mojito, Highball glass
\end{minipage}
\hspace{0.1cm}
\begin{minipage}{0.45\textwidth}
    \NegativeFewShot Smoothie, Tea cup
    
    \NegativeFewShot Iced Cola, Glass
    
    \NegativeFewShot Orange Juice, Juice Glass
    
    \NegativeFewShot Hot Coffee, Coffee Mug
    
    \NegativeFewShot Beer, Beer Mug
    
    \NegativeFewShot Lemonade, Pitcher
    
    \NegativeFewShot Water, Water Bottle
    
    \NegativeFewShot Wine, Wine Glass
    
    \NegativeFewShot Iced Mocha, Soda Can
    
    \NegativeFewShot Cold Brew, Wine Glass
    
    \NegativeFewShot Mojito, Pint Glass
    
    \NegativeFewShot Cider, Champagne Flute
    
    \NegativeFewShot Iced Milk Latte, Pint Glass
\end{minipage}

\vspace{0.1cm}
\small
{\bf Answer:} Sure! Based on the examples you provided, I will generate 30 new positive noun-concept pairs (beverage, uncommon vessel). I will ensure that the nouns are not repeated. Here are the pairs:
\vspace{1mm}

\scriptsize
\begin{minipage}{0.45\textwidth}
    1. \PositiveFewShot Ginger Ale, Martini Glass
    
    2. \PositiveFewShot Root Beer, Teapot
    
    3. \PositiveFewShot Soda, Whiskey Flask
    
    4. \PositiveFewShot Chocolate Milk, Wine Carafe
    
    5. \PositiveFewShot Matcha Latte, Beer Boot

    6. \PositiveFewShot Bloody Mary, Coffee Carafe
    
    7. \PositiveFewShot Seltzer, Whiskey Decanter
    
    8. \PositiveFewShot Lemonade, Margarita Glass
    
    9. \PositiveFewShot Sprite, Tea cup
    
    10. \PositiveFewShot Lassi, Martini Shaker
    
    11. \PositiveFewShot Fruit Punch, Whiskey Flask
    
    12. \PositiveFewShot White Wine, Shot Glass
    
    13. \PositiveFewShot Champagne, Milk Jug
    
    14. \PositiveFewShot Red Wine, Milkshake Glass
    
    15. \PositiveFewShot Diet Coke, Teapot
    
\end{minipage}
\hspace{0.5cm}
\begin{minipage}{0.5\textwidth}
    16. \PositiveFewShot Pepsi, Champagne Bucket

    17. \PositiveFewShot Mountain Dew, Wine Decanter
    
    18. \PositiveFewShot Black Tea, Margarita Glass
    
    19. \PositiveFewShot Mint Julep, Coffee Carafe
    
    20. \PositiveFewShot Rum, Tea Kettle
    
    21. \PositiveFewShot Whiskey, Juice Box

    22. \PositiveFewShot Vodka, Soup Bowl
    
    23. \PositiveFewShot Gin and Tonic, Coffee Thermos
    
    24. \PositiveFewShot Sprite, Wine Bucket
    
    25. \PositiveFewShot Protein Shake, Wine Carafe
    
    26. \PositiveFewShot Kombucha, Martini Glass
    
    27. \PositiveFewShot Grey Tea, Champagne Bucket
    
    28. \PositiveFewShot Cappuccino, Margarita Glass
    
    29. \PositiveFewShot Fanta, Coffee Pot
    
    30. \PositiveFewShot Tonic Water, Teapot
    
\end{minipage}

\end{tcolorbox}

\newpage

\textbf{Phase 3 - Discovering New Patterns for Concept Pair Generation}
The proliferation of repetitive concept pairs has prompted researchers to investigate innovative approaches. Engaging with \gpt, scholars aim to automate the discovery of new patterns. Our instructions for \gpt are as follows:

\begin{tcolorbox}[colback=blue!5!white,colframe=blue!75!black, before skip=5pt]
            \small {\bf Prompt:} I want you to act as a bot summarizing the pattern rule and generating more patterns. You are good at text-image concept grounding, visual concept understanding, text-to-image diffusion models' prompt engineering. Don't be verbose in your answer, but do provide details and examples that  might help the explanation. 

            \small
            I have provided you with a series of existing positive and negative patterns, where "pattern" refers to a generalization of certain noun pairs, A and B.  Where A and B are very common concepts, such as coke and tea cup. Moreover, these noun pairs are closely related to visual imagery, rather than being two independent objects, such as the wrapper and the wrapped item, background and foreground, etc.

            In the provided positive and negative patterns, the noun pairs in the positive pattern are combinations that almost never appear in the text-to-image training dataset, such as beverage and erroneous container (e.g., iced cola and tea cup) or local cuisine and non-native location (e.g., sushi and New York City). Specifically, for a positive noun pair "A" and "B", there will usually be a hidden concept "C". "A" always appears with "C" in the text-to-image training dataset, "B" and "C" belong to the same category or are of the same kind but their visual effect is very different. Therefore, because of the strong association between "A" and "C," it leads to "B" being obscured or occupied in the generated image with prompts "A" and "B". For example, in the positive pattern "Beverage and erroneous container", "A" (iced coke) and "B" (tea cup) are a good example pair. Because "A" (iced coke) is strongly bound to the latent concept "C" (glass), telling the model to generate "a tea cup of iced coke" will generate a glass of iced coke instead of a tea cup of iced coke.

            Conversely, the noun pairs in the negative pattern are combinations that frequently appear in the text-to-image training dataset, such as animal and their corresponding external covering or jackets and their appropriate inner garments.
            Your task is to come up with some new positive patterns based on the given positive and negative patterns:

            (see next page)

\end{tcolorbox}

\begin{tcolorbox}[colback=blue!5!white,colframe=blue!75!black]

            {\small
            \vspace{2mm}
            
            \PositiveFewShot  Beverage and erroneous container
            
            \PositiveFewShot  Animal and incorrect external covering
            
            \PositiveFewShot  Mismatched outfits
            
            \PositiveFewShot  Endangered species and historical sites
            
            \PositiveFewShot  Local cuisine and non-native location
            
            \NegativeFewShot  Food and its appropriate container
            
            \NegativeFewShot  Furniture and its corresponding setting
            
            \NegativeFewShot  Endangered animals and their habitats.
            }

            \vspace{0.2cm}
            {\bf Answer:} Certainly, here are some new positive patterns based on the provided guidelines:

            \vspace{2mm}
            1. \PositiveFewShot Musical Instrument and Inappropriate venue
            \\- Example: Violin and Football stadium
            \\- Explanation: Visualize a violin being in a football stadium, creating a striking visual contrast between the delicate instrument and the sports arena.
            
            2. \PositiveFewShot Fruit and Unrelated seasoning
            \\- Example: Orange and Salt shaker
            \\- Explanation: Imagine an orange paired with a salt shaker, where the seasoning doesn't match the typical use for the fruit.

            \small
            
            3. \PositiveFewShot Spacecraft and Terrestrial landscape
            \\- Example: Rocket and Desert landscape
            \\- Explanation: Picture a rocket placed in a desert landscape, highlighting the visual discord between space technology and earthly terrain.
            
            4. \PositiveFewShot Jewelry and Improper storage
            \\- Example: Diamond necklace and Toolbox
            \\- Explanation: Visualize a valuable diamond necklace stored in a toolbox, emphasizing the incongruity between luxury jewelry and a utilitarian container.
            
            5. \PositiveFewShot Sports Gear and Wrong sport
            \\- Example: Baseball bat and Ice hockey rink
            \\- Explanation: Picture a baseball bat in an ice hockey rink, where the sporting equipment is out of place for the chosen location.
            \\......
            \vspace{2mm}
            These patterns create visual contradictions by juxtaposing related concepts in unexpected ways, following the same principles as the given positive patterns.
\end{tcolorbox}

Due to the lack of guidance from information that has been validated in visual space, the new concept pairs answered by \gpt don't achieve the expected effect.
So human researchers propose a shift toward necessitating concepts within the linguistic domain rather than demanding concept pairs specifically.
In this way, human researchers only need to select suitable conflicting concepts from two sets of concepts and combine them for verification, drawing from their accumulated experience with text-image interactions.
This avoids both unguided exploration and excessive guidance, see the instruction prompt in the next page. 

\newpage

\begin{tcolorbox}[colback=blue!5!white,colframe=blue!75!black]
            \small {\bf Prompt:} What storage is the most unlikely to put jewelry like a pearl ring necklace in? Help me list some of the most unlikely.
            \\
            \\
            {\bf Answer:} When considering containers that are highly unlikely choices for storing a precious item like a pearl ring necklace, we can think creatively. Here are some of the most unlikely containers:

            \vspace{2mm}
            1. Bird's Nest: Placing a delicate pearl ring necklace in a bird's nest would be highly impractical and unexpected.
            
            2. Fish Tank: The underwater environment of a fish tank would be detrimental to a pearl ring necklace.
            
            3. Toilet Bowl: Using a toilet bowl as a storage place for jewelry is not only unhygienic but also absurd.
            
            4. Compost Bin: Pearls and organic waste do not mix, making a compost bin a truly improbable choice.
            
            5. Dog's Food Bowl: Storing jewelry alongside pet food is both unclean and unusual.

            \small
            6. Fireplace Ashes: The ashes from a fireplace are the last place one would think of storing a delicate necklace.
            
            7. Shoe Sole: Tucking a pearl necklace into the sole of a shoe is an unexpected and risky choice.
            
            8. Fishing Tackle Box: A tackle box designed for fishing equipment is far from suitable for jewelry storage.
            
            9. Hollowed-Out Watermelon: Hiding jewelry inside a watermelon would be a whimsical and impractical choice.
            
            10. Tool Shed Toolbox: Storing jewelry in a toolbox designed for tools is unconventional and risky.

            \vspace{2mm}
            These containers not only provide an answer to your question but also illustrate the unusual and impractical nature of storing a precious item like a pearl ring necklace in them. 
\end{tcolorbox}

We have also present the new patterns given by \gpt in Tabel~\ref{tab:new_pattern} as follows:

\begin{table}[h]
\caption{New patterns created by \gpt and the levels of their concept pairs.}
\label{tab:new_pattern}
\centering
\begin{tabular}{ccc}
\toprule
Category                        & New Pattern                                & \begin{tabular}[c]{@{}c@{}}Num of Pairs\\ Rated at Level 5\end{tabular} \\ \midrule
\multirow{6}{*}{\ourProblem} & Jewelry and improper storage               & 11 / 15                                                                 \\
                                & Flower and unconventional vase             & 10 / 15                                                                 \\
                                &  Sport and unconventional equipment                  &  13 / 15                                                                       \\
                                & Musical instrument and inappropriate venue & 11 / 15                                                                 \\
                                & Sports equipment and mismatched locations   & 10 / 15                                                                  \\
                                & Household appliances and incorrect room & 8 / 15                                                                     \\ \bottomrule
\end{tabular}
\end{table}

\newpage

\textbf{Phase 4 - Creating Novel Concept Pairs by Merging Patterns}
Two straight lines determine a plane.
Once the points on the two lines are defined, it's available to guide \gpt to expand the new plane. This means \gpt can blend newly generated patterns with previous patterns, thereby generating more \ourProblem concept pairs. Our instructions for \gpt are as follows:
\begin{tcolorbox}[colback=blue!5!white,colframe=blue!75!black, before skip=5pt]
            \small {\bf Prompt:} I found that the text-to-image model cannot draw some images specified by patterns. For example, the pattern "Beverage and wrong beverage container" ("A and B") and the pattern "Jewelry and Impre storage" ("C and D").
            
            The pattern "Beverage and erroneous container" includes many concept pairs, such as "iced coke in a tea cup", "green tea in a red solo cup", "Orange juice in a coffee mug", and so on.
            
            And another pattern "Jewelry and improper storage" also includes many concept pairs, such as "Diamond necklace in toolbox", "Ruby earings in bird nest", "Silver anklet in toilet bowl", and so on.
            
            I hope to come up with more patterns that cannot be drawn based on these two patterns. Can you help me come up with a few more patterns based on these two patterns. In other words, I mean you to generate new patterns by recombining the items in the two patterns I provided to you, rather than providing new concept pairs in the patterns I gave. You need to use only the two patterns I've provided and regroup things.
            
            In the current case, "A" refers to beverage, "B" refers to beverage containers, "C" refers to Jewelry, and "D" refers to storage. Give me the appropriate 2 patterns "A and D" and "B and C". Where "A" and "D" almost never occur together, and "B" and "C" never occur together. For each new pattern, please give me detailed concept pairs belonging to this pattern that the model may fail to draw such an image.

            {\bf Answer:} Certainly, I can help you generate two new patterns "A and D" and "B and C" based on your provided patterns "Jewelry and improper storage" and "Beverage and erroneous container" These new patterns will involve combinations that are less likely to occur together, and the model may have difficulty drawing corresponding images.

            \vspace{2mm}
            Pattern 1: "A and D" - Beverage and Imperial Storage
            
            In this pattern, we combine "Beverage" (A) with "Imperial storage" (D). These two elements are quite different, and the challenge here would be to depict beverages in an imperial storage setting.
            
            1. Champagne in a treasure chest
            
            2. Coffee in a fish tank
            
            3. Soda in a bird nest
            
            4. Tea in an ornate jewelry box
            
            5. Red wine in a toolbox

            ......
            
            (see next page)
\end{tcolorbox}

\begin{tcolorbox}[colback=blue!5!white,colframe=blue!75!black]
            Pattern 2: "B and C" - Beverage Container and Jewelry
            
            This pattern combines "Beverage container" (B) with "Jewelry" (C). These elements typically have no connection, so the challenge would be to depict jewelry inside beverage containers.
            
            1. Diamond ring in a soda can
            
            2. Gold necklace in a coffee thermos
            
            3. Pearl bracelet in a beer mug
            
            4. Sapphire earrings in a wine bottle
            
            5. Ruby brooch in a cocktail shaker

            ......

            \vspace{2mm}
            These new patterns should provide a creative challenge for the text-to-image model by combining elements that are not usually associated with each other.
\end{tcolorbox}

Human researchers then repeat the verification process used in Phase 1 to verify these concept pairs.
After thorough verification, we report the number of concept pairs rated as Level 5 for each blended pattern in Table~\ref{tab:blend_pattern}.
It's quite surprising to discover that the validation results closely align with the expectations of human researchers, indicating that \gpt successfully integrates two mutually orthogonal patterns.
\begin{table}[h]
\caption{Blended patterns created by \gpt and the levels of their concept pairs.}
\label{tab:blend_pattern}
\centering
\begin{tabular}{ccc}
\toprule
Category                         & Blended Pattern                         & \begin{tabular}[c]{@{}c@{}}Num of Pairs\\ Rated at Level 5\end{tabular} \\ \midrule
\multirow{12}{*}{\ourProblem} & Jewelry and beverage container          & 12 / 15                                                                 \\
                                 & Beverage and jewelry storage            & 10 / 15                                                                 \\
                                 & Flower and beverage container           & 10 / 15                                                                 \\
                                 & Beverage and flower vase                & 11 / 15                                                                 \\
                                 & Sport and wrong beverage container                 &  15 / 15                                                                      \\
                                 &  Unconventional equipment and beverage                                       &  12 / 15                                                                       \\
                                 & Musical instrument and historical sites & 10 / 15                                                                 \\
                                 & Endangered species and musical scenes   & 9 / 15                                                                  \\
                                 &   Sports equipment in the wrong type of room                                       &  10 / 15                                                                          \\
                                 &         Locations with misplaced household appliances                               &                12 / 15                          \\
                                 & Household appliances and location outside of origin & 11 / 15 \\

                                 & Incorrect room and local food & 12 / 15 \\
 \bottomrule
\end{tabular}
\end{table}

\newpage
\section{Interaction in Sequential Concept Introduction}
Here, we present the detailed interaction to guide \gpt to provide the most logical sequence of two concepts and the description of them.

\begin{tcolorbox}[colback=blue!5!white,colframe=blue!75!black]
            \small {\bf Prompt:}  I want you to act as an AI creative assistant. I have two concepts A and B, and a sentence combining these two concepts together. I want you to help me generate a  brief visual description of each concept. Namely, a concise description of what a given noun itself looks like, such as color, shape, or other visual features. As short as possible. No background. No other objects.
            
            In addition, if I tend to draw a picture using the sentence, please tell me which concept I should draw first. For example, if a concept is a type of container, we should draw such a container first. If a concept is a type of background, we should draw such background first.

            \small
            Now, the concept A is "a tea cup", B is "iced coke", and the sentence is "A tea cup filled with iced coke". Please answer me with the descriptions of A and B, and tell me which one to draw first.
            \\
            \\
            {\bf Answer:}
            Certainly, I can help you. 
            
            Concept A: Tea cup
            - A small, usually white or pastel-colored cup, often with a handle, designed to hold hot beverages.

            Concept B: Iced Coke
            - Dark, fizzy liquid with translucent ice cubes floating within.
            \vspace{2mm}
            \\To draw the sentence "A tea cup filled with iced coke":
            \\1. Draw the tea cup first (Concept A), as it acts as the container.
            \\2. Then, fill the tea cup with the iced coke (Concept B).
\end{tcolorbox}

\section{Comprehensive Explanation of MoCE}
\label{sec:sup_moce}
\begin{figure}[h]
    \centering
    \includegraphics[width=\linewidth]{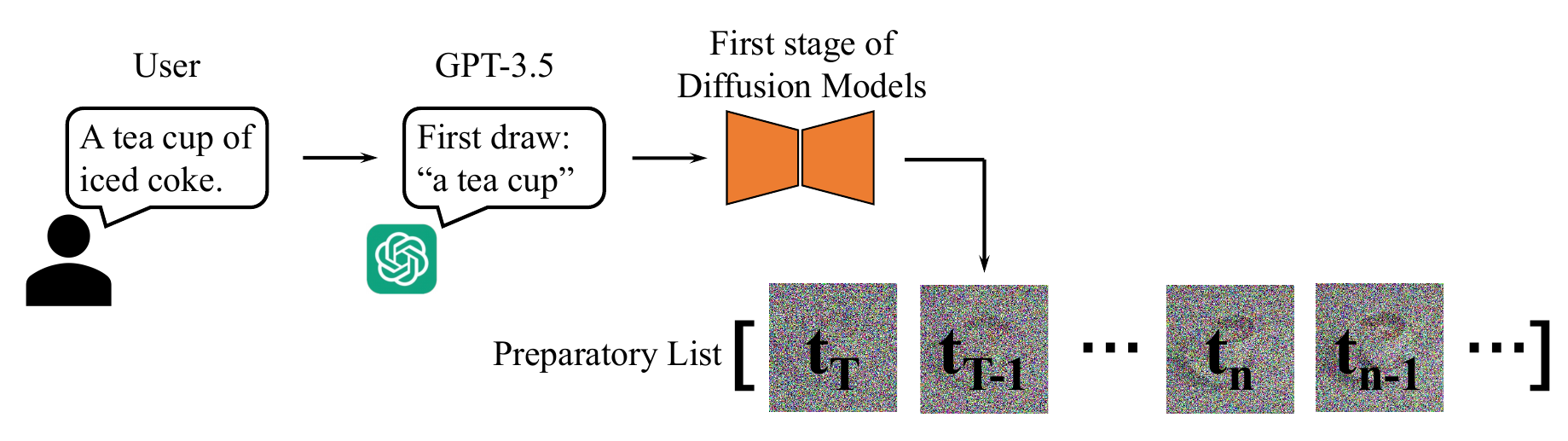}
    \caption{First phase of \ourMethod. Provided with a prompt, \gpt determines drawing order. \sdxl generates latent images at each step, storing them in a preparatory list.}
    \label{fig:method1}
\end{figure}

\begin{figure}[h]
    \centering
    \includegraphics[width=\linewidth]{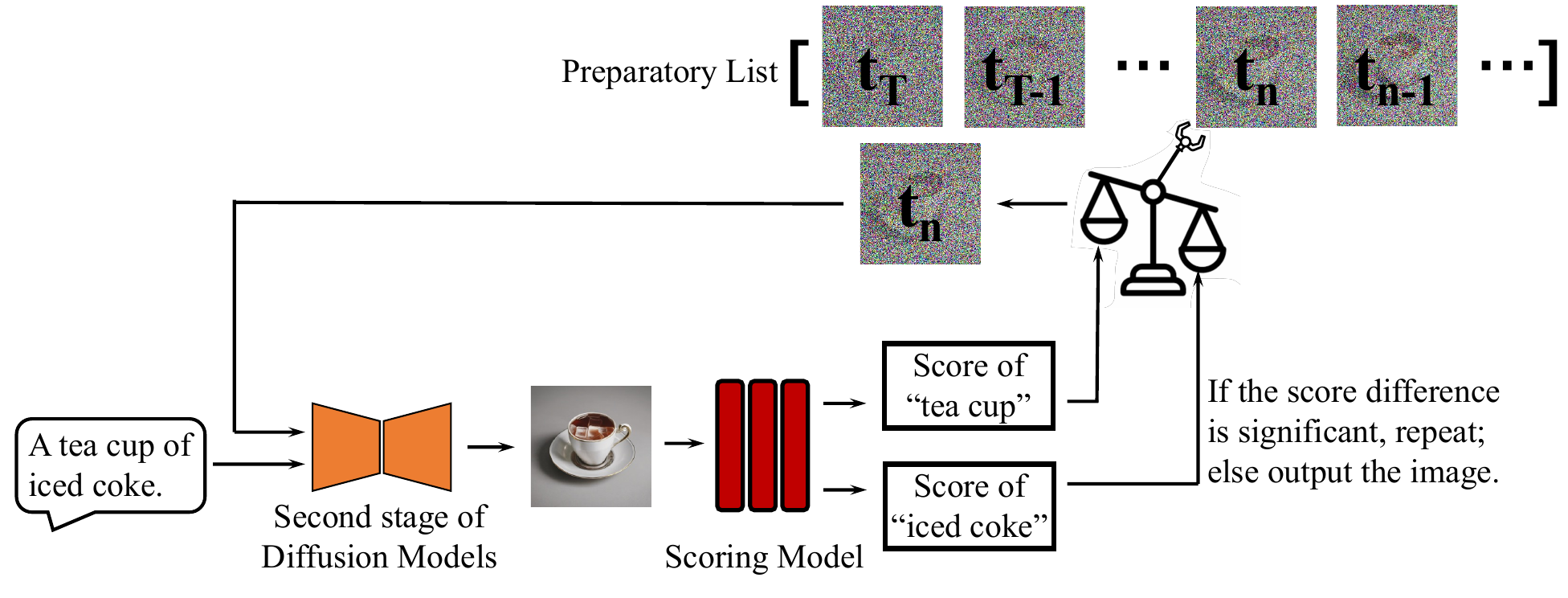}
    \caption{Second phase of \ourMethod. The second stage model takes an image from the preparatory list and performs denoising. A specific selection process, utilizing the \ourMetric and binary search, was used to select a better image from the list.}
    \label{fig:method2}
\end{figure}

The whole process of \ourMethod actually contains two phases. For the first phase, as depicted in Figure \ref{fig:method1}, Diffusion Models was asked to generate the image for the selected substance to form a preparatory list including images of different timestamps $t_T$.
From there, the second phase of \ourMethod, as shown in Figure \ref{fig:method2} gets in effect, which takes an image of specific timestamp $t_n$ from the preparatory list as the input for the Diffusion Models in the second stage where the whole sentence containing both substances was taken as the prompt. The selection of the specific candidate from the preparatory list was determined by our scoring model introduced in paper Section~\ref{sec:our_method}.

\section{Score Evaluation}
\label{sec:sup_score_eval}
By convention, evaluation metrics such as \clipscore and \imgreward are usually used to gauge the correspondence between the generations and the input entities quantitatively. 
These metrics utilize the cosine similarity between embeddings produced by Deep Neural Networks (DNNs).
Nonetheless, they often fail to discern numerical values, transparent objects, and other crucial elements readily identifiable by humans.
In our paper Section~\ref{sec:our_method}, our proposed \dynamic uses the Multi-Concept Disparity ($\mathcal{D}$) between the scores of 64 images ($M$) with respect to 2 concepts (\conceptA and \conceptB) as the basis for performing a binary search:
\begin{equation}
    \label{eq:score_difference_in_section_exp}
    \mathcal{D} = \mathcal{S}(M, \mathcal{A}) - \mathcal{S}(M, \mathcal{B})
\end{equation}
We use this metric to assist the demonstration our \dynamic performance. Specifically, we assess the generation performance of both the baseline model (\sdxl) and our \dynamic using the metric, $\mathcal{D}$, in Equation~\ref{eq:score_difference_in_section_exp}, where $\mathcal{D}$ is calculated using \clipscore or \imgreward, denoted as $\mathcal{D}$ - \clipscore\footnote{To facilitate easier observation by human experts, we demonstrate the established $\mathcal{D}$ - \clipscore at a magnification of $10\times$.} and $\mathcal{D}$ - \imgreward respectively.
Experiments are conducted on both the set of concept pairs at Level 5 and Level 1 to 4.
We report the experimental results in Table~\ref{tab:score_eval}.
In comparison to the baseline model, $\mathcal{D}$ - \clipscore of images generated by our \dynamic is reduced by more than half, and the $\mathcal{D}$ - \imgreward is reduced by more than $\frac{1}{3}$, in either set of levels.
It demonstrates \dynamic's ability to effectively restore the lost concepts in images.
\begin{table}[h]
\caption{Score Evaluation for our \dynamic using both $\mathcal{D}$ - \clipscore ($\downarrow$) and $\mathcal{D}$ - \imgreward ($\downarrow$). We use concept pairs originally rated as Level 5 and Level 1 - 4.}

\label{tab:score_eval}
\centering
\begin{tabular}{cccc}
\toprule
Original Level               & Method                  & $\mathcal{D}$ - \clipscore ($\downarrow$) & \quad $\mathcal{D}$ - \imgreward ($\downarrow$) \\ \midrule 
\multirow{2}{*}{Level 1 - 4} & Baseline                & 0.91                                                   & 1.10                                                   \\ 
                             & \dynamic & \textbf{0.45}                                          & \textbf{0.75}                                          \\ \midrule 
\multirow{2}{*}{Level 5}     & Baseline                & 1.09                                                   & 1.21                                                   \\
                             & \dynamic & \textbf{0.55}                                          & \textbf{0.79}                                          \\ \bottomrule
\end{tabular}
\end{table}

\section{Ablation Demo}
\begin{figure}[h]
        \centering
        \includegraphics[width=0.5\linewidth]{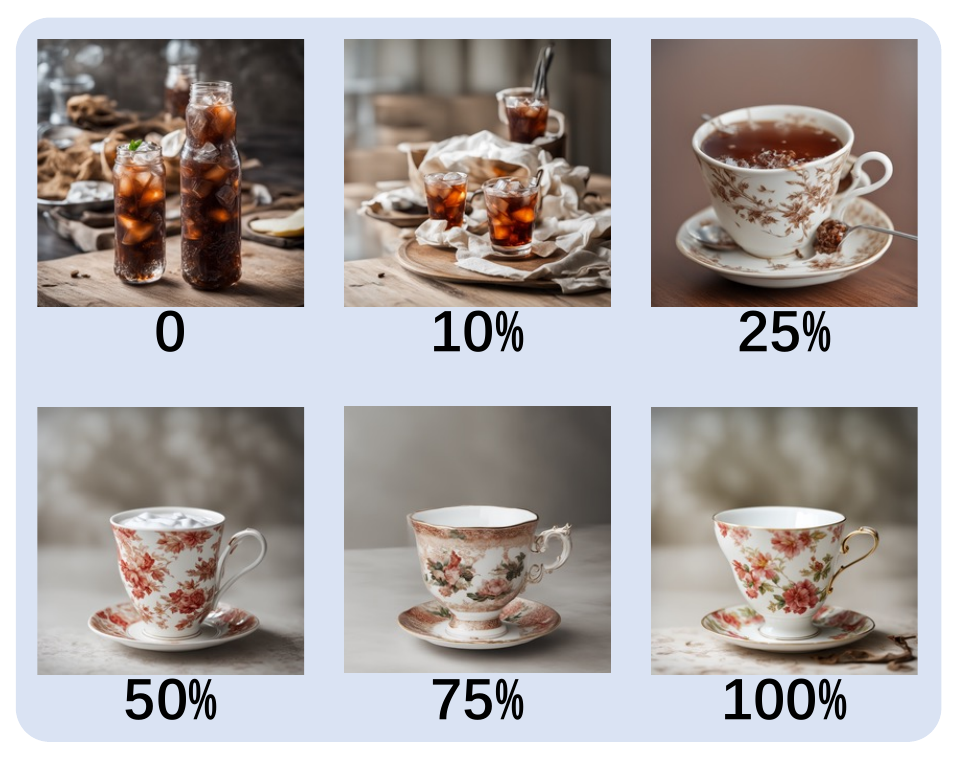}
        \captionof{figure}{Analysis for different prompt switching times on ``a tea cup of iced coke."}
        \label{figure:time_ablation}
\end{figure}

Here, we first demonstrate the ablation test on computational time. Figure \ref{figure:time_ablation} shows that the prompt switching time can be set between 20\% and 40\% of the steps to optimize the time-performance tradeoff, thus optimizing the cost to 1$\times$ or 2$\times$ at the same time.

\begin{figure}[htbp]
        \centering
        \includegraphics[width=\linewidth]{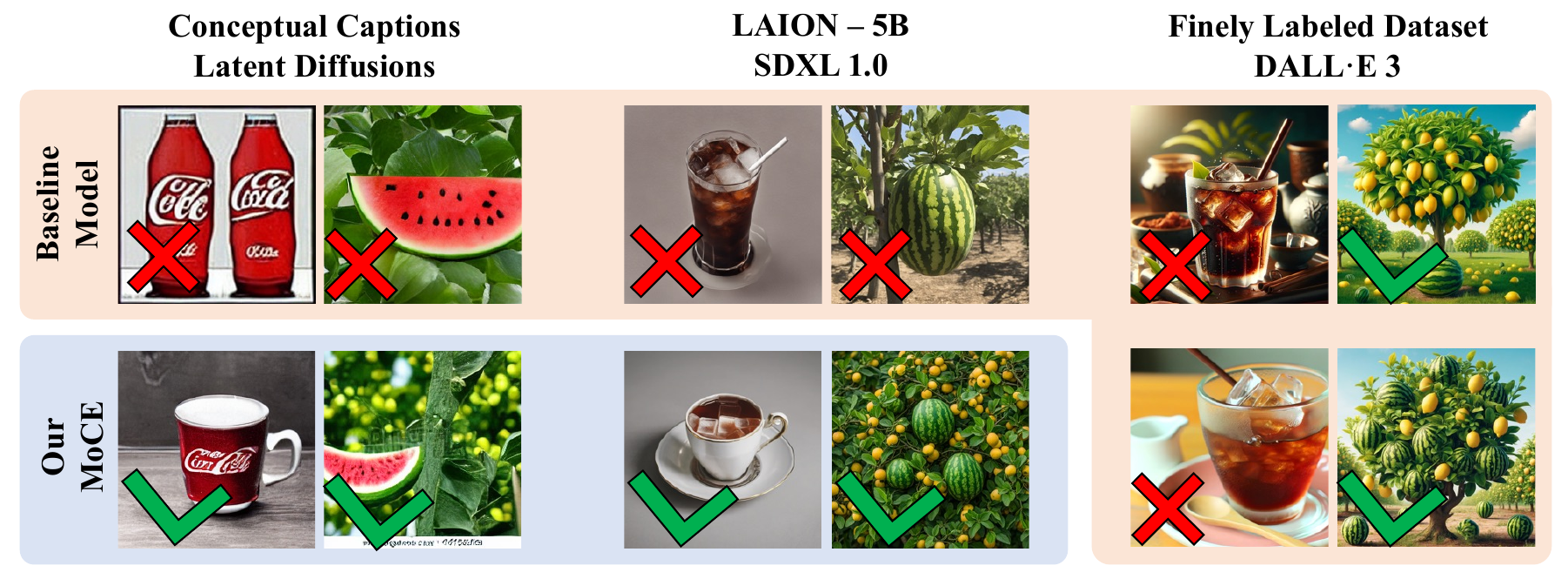} 
        \captionof{figure}{Comparison between baseline models and MoCE on various datasets with different sizes.}
        \label{figure:dataset_ablation}
\end{figure}

Moreover, ablation test on dataset size is also conducted. Here we compare 3 models from datasets of varying sizes, as shown in Figure \ref{figure:dataset_ablation}. Since DALL·E 3 is not open-sourced, we cannot apply MoCE to it. DALL·E 3 does perform well. However, its detailed labeling process is labor-intensive,while at the same time, our MoCE can be easily integrated into current models.

\section{Restoration Visualizations of Level 5}
\begin{figure*}[htbp!]
    \begin{center}
        \adjustbox{max width=1\linewidth}{
            \begin{tabular}{c|cccc}
            \resizebox{!}{12mm}{Baseline Models} & \multicolumn{4}{c}{\resizebox{!}{12mm}{\dynamic}} \\
            \includegraphics[width=1\linewidth]{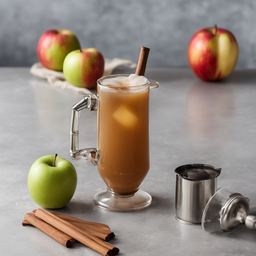} &
            \includegraphics[width=1\linewidth]{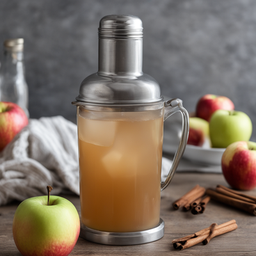} &
            \includegraphics[width=1\linewidth]{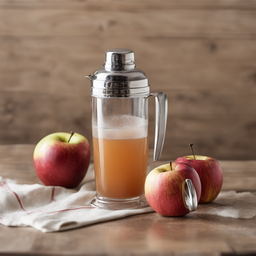} &
            \includegraphics[width=1\linewidth]{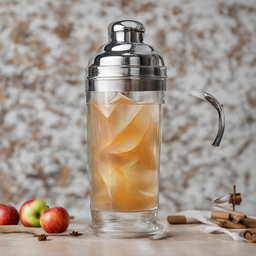} &
            \includegraphics[width=1\linewidth]{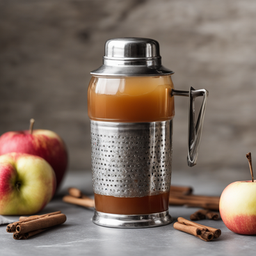} \\
             & \multicolumn{3}{c}{\resizebox{!}{12mm}{(a) Apple Cider, Cocktail Shaker}} & \\
            \includegraphics[width=1\linewidth]{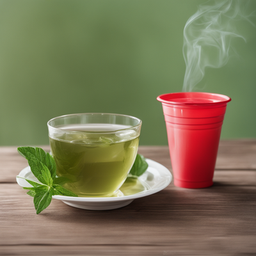} &
            \includegraphics[width=1\linewidth]{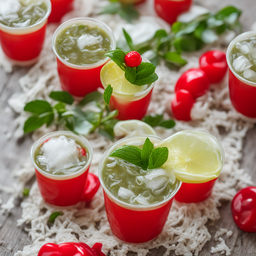} &
            \includegraphics[width=1\linewidth]{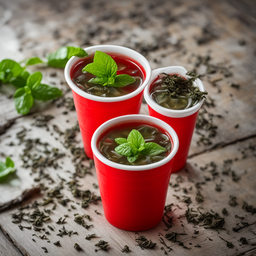} &
            \includegraphics[width=1\linewidth]{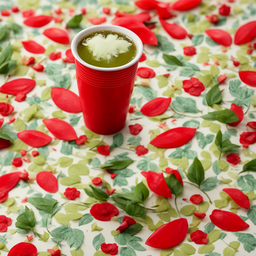} &
            \includegraphics[width=1\linewidth]{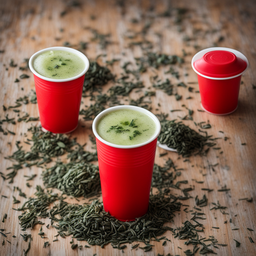} \\
             & \multicolumn{3}{c}{\resizebox{!}{12mm}{(b) Green Tea, Red Solo Cup}} & \\
            \includegraphics[width=1\linewidth]{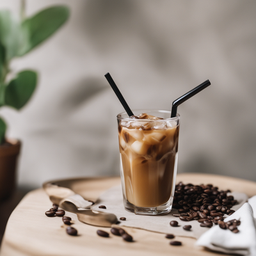} &
            \includegraphics[width=1\linewidth]{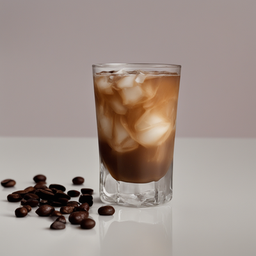} &
            \includegraphics[width=1\linewidth]{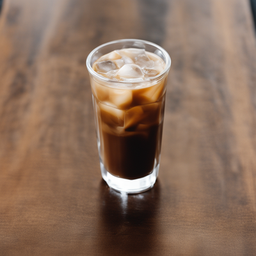} &
            \includegraphics[width=1\linewidth]{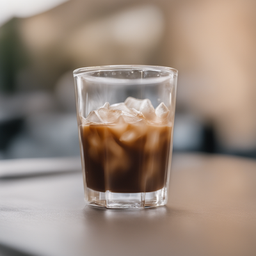} &
            \includegraphics[width=1\linewidth]{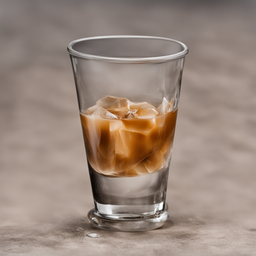} \\
             & \multicolumn{3}{c}{\resizebox{!}{12mm}{(c) Iced Coffee, Shot Glass}} 
             & \\
            \includegraphics[width=1\linewidth]{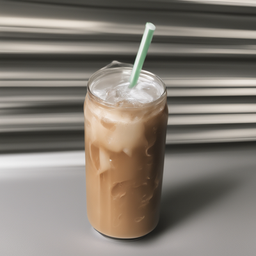} &
            \includegraphics[width=1\linewidth]{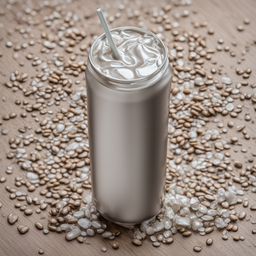} &
            \includegraphics[width=1\linewidth]{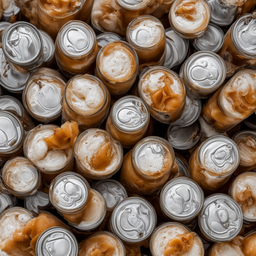} &
            \includegraphics[width=1\linewidth]{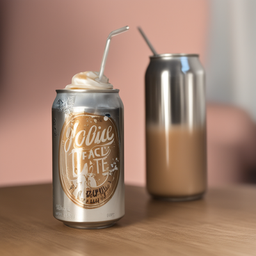} &
            \includegraphics[width=1\linewidth]{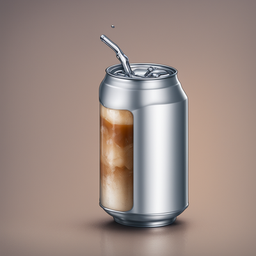} \\
            & \multicolumn{3}{c}{\resizebox{!}{12mm}{(d) Iced Latte, Soda Can}} &\\
            \includegraphics[width=1\linewidth]{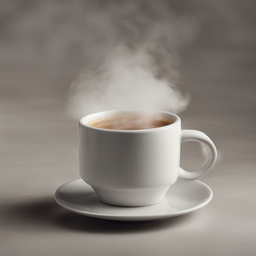} &
            \includegraphics[width=1\linewidth]{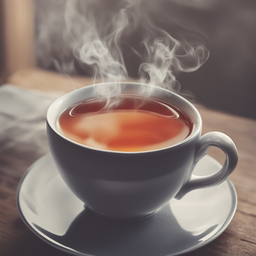} &
            \includegraphics[width=1\linewidth]{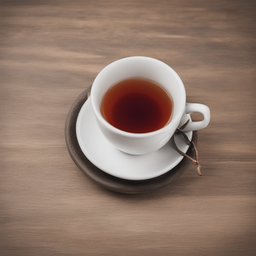} &
            \includegraphics[width=1\linewidth]{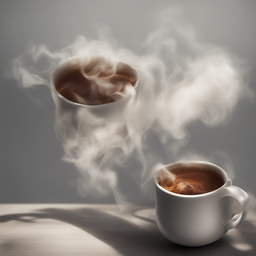} &
            \includegraphics[width=1\linewidth]{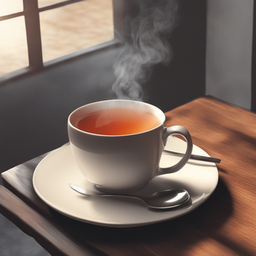} \\
            & \multicolumn{3}{c}{\resizebox{!}{12mm}{(e) Hot Tea, Coffee Mug}} &\\
            \end{tabular}
            }
        \end{center}
    \caption{Visualizations of images at Level 5 generated by baseline models and our \dynamic.}
    \label{fig:level5_fix_a}
\end{figure*}

\begin{figure*}[t]
    \begin{center}
        \adjustbox{max width=1\linewidth}{
            \begin{tabular}{c|cccc}
            \resizebox{!}{12mm}{Baseline Models} & \multicolumn{4}{c}{\resizebox{!}{12mm}{\dynamic}} \\
            \includegraphics[width=1\linewidth]{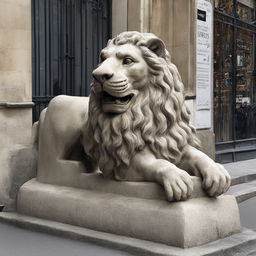} &
            \includegraphics[width=1\linewidth]{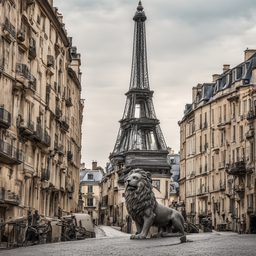} &
            \includegraphics[width=1\linewidth]{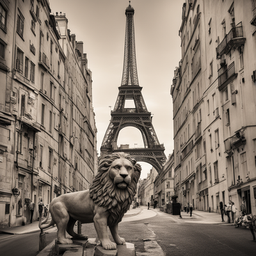} &
            \includegraphics[width=1\linewidth]{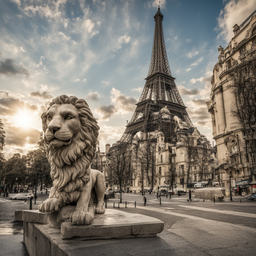} &
            \includegraphics[width=1\linewidth]{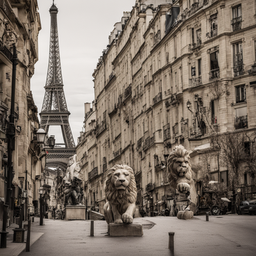} \\
            & \multicolumn{3}{c}{\resizebox{!}{12mm}{(f) Stone Lion, Paris}} & \\
            \includegraphics[width=1\linewidth]{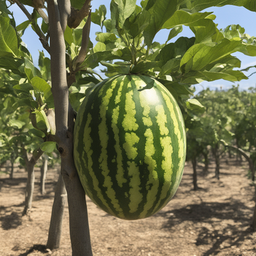} &
            \includegraphics[width=1\linewidth]{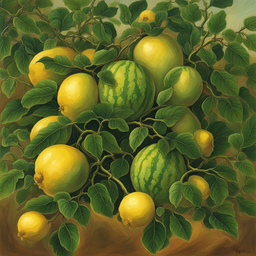} &
            \includegraphics[width=1\linewidth]{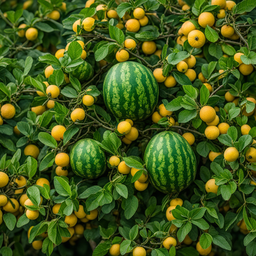} &
            \includegraphics[width=1\linewidth]{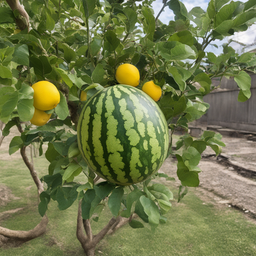} &
            \includegraphics[width=1\linewidth]{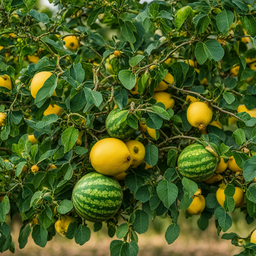} \\
            & \multicolumn{3}{c}{\resizebox{!}{12mm}{(g) Watermelon, Lemon Tree}} & \\
            \includegraphics[width=1\linewidth]{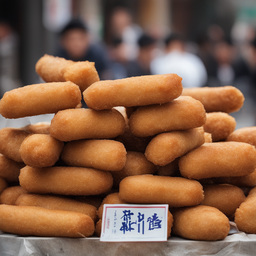} &
            \includegraphics[width=1\linewidth]{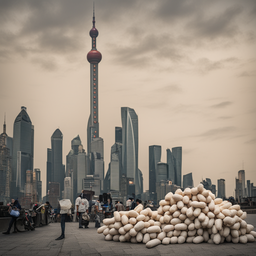} &
            \includegraphics[width=1\linewidth]{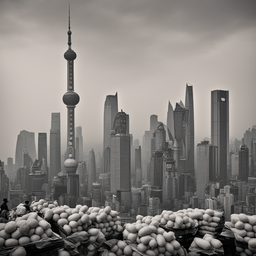} &
            \includegraphics[width=1\linewidth]{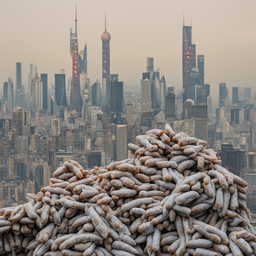} &
            \includegraphics[width=1\linewidth]{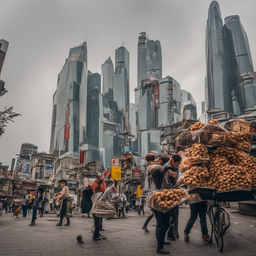} \\
             & \multicolumn{3}{c}{\resizebox{!}{12mm}{(h) Croquettes, Shanghai}} & \\
            \includegraphics[width=1\linewidth]{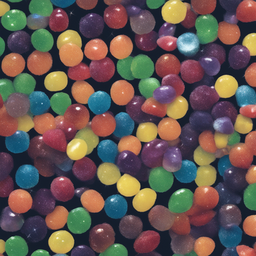} &
            \includegraphics[width=1\linewidth]{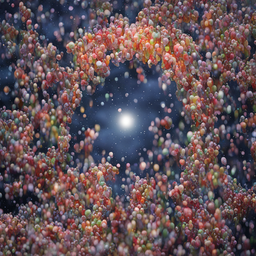} &
            \includegraphics[width=1\linewidth]{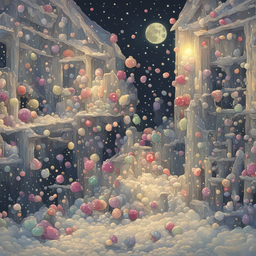} &
            \includegraphics[width=1\linewidth]{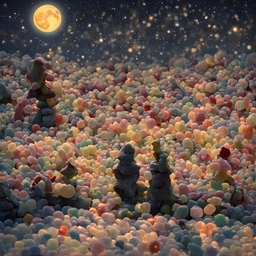} &
            \includegraphics[width=1\linewidth]{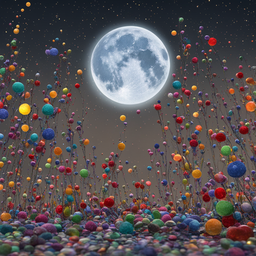} \\
             & \multicolumn{3}{c}{\resizebox{!}{12mm}{(i) Gumdrops, Night}} & \\
            \includegraphics[width=1\linewidth]{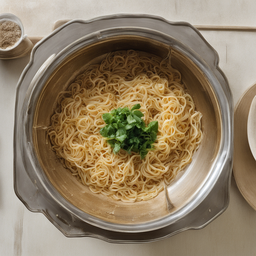} &
            \includegraphics[width=1\linewidth]{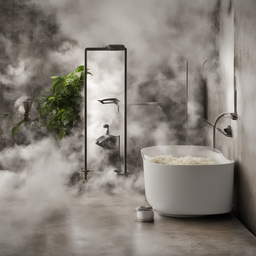} &
            \includegraphics[width=1\linewidth]{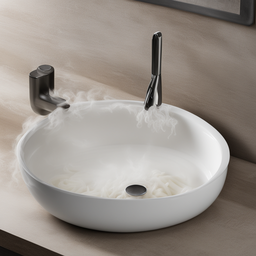} &
            \includegraphics[width=1\linewidth]{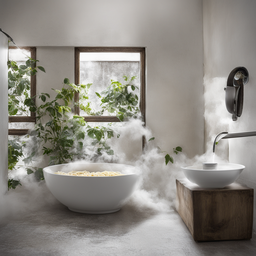} &
            \includegraphics[width=1\linewidth]{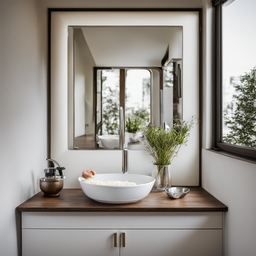} \\
            & \multicolumn{3}{c}{\resizebox{!}{12mm}{(j) Noodle, Basin}} & \\
            \end{tabular}
        }
    \end{center}
    \caption{Visualizations of images at Level 5 generated by baseline models and our \dynamic.}
    \label{fig:level5_fix_b}
\end{figure*}

Here, we demonstrate more visualizations of images of Level 5 restored using our \dynamic.
In spite of the given additional rich information, \mj fails to correctly generate these images.
While using the same text prompts, our \dynamic successfully retrieves the lost concepts as presented in Figure~\ref{fig:level5_fix_a} and~\ref{fig:level5_fix_b}.
We also propose that in Figure \ref{fig:level5_fix_a} and \ref{fig:level5_fix_b}, human experts judge “Hot Tea” based on the transparency of the liquid and “Shanghai” based on landmarks such as the Oriental Pearl TV Tower. These judgments are based on verified model tendencies.

\section{Restoration Visualizations of Level 1 - 4}
\begin{figure*}[h]
    \begin{center}
        \begin{tikzpicture}
            \node[draw=none,fill=none,text=black,font=\fontsize{10}{10}\selectfont] at (0.25\linewidth, 0) {Baseline Models};
            \node[draw=none,fill=none,text=black,font=\fontsize{10}{10}\selectfont] at (0.5\linewidth, 0.4) {Baseline Models};
            \node[draw=none,fill=none,text=black,font=\fontsize{10}{10}\selectfont] at (0.5\linewidth, 0) {with Additional Info};
            \node[draw=none,fill=none,text=black,font=\fontsize{10}{10}\selectfont] at (0.74\linewidth, 0) {\dynamic};
        \end{tikzpicture}
        \adjustbox{max width=\linewidth}{
            \begin{tabular}{c c c}
            \includegraphics[height=2.8cm]{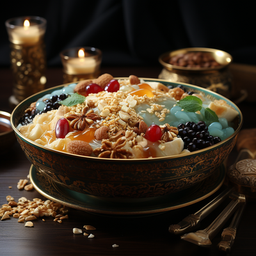} &
            \includegraphics[height=2.8cm]{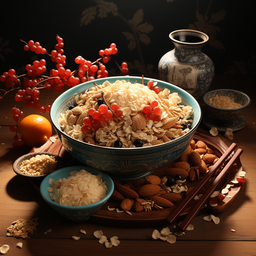} &
            \includegraphics[height=2.8cm]{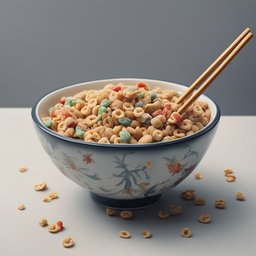} \\
             \multicolumn{3}{c}{(a) Cereal, Chopsticks} \\
            \includegraphics[height=2.8cm]{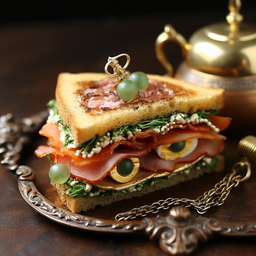} &
            \includegraphics[height=2.8cm]{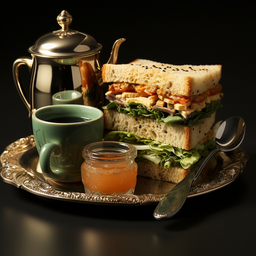} &
            \includegraphics[height=2.8cm]{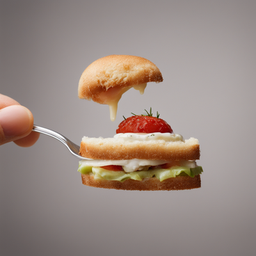} \\
             \multicolumn{3}{c}{(b) Teaspoon, Sandwich} \\
            \includegraphics[height=2.8cm]{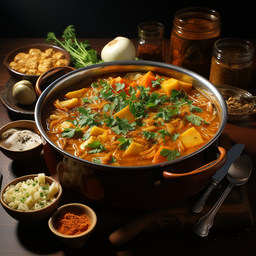} &
            \includegraphics[height=2.8cm]{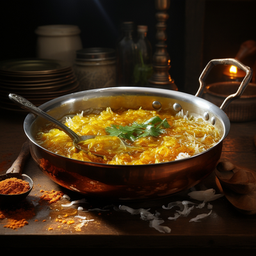} &
            \includegraphics[height=2.8cm]{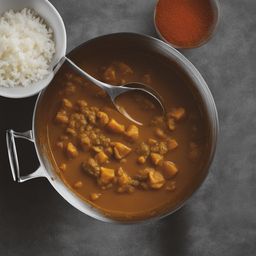} \\
             \multicolumn{3}{c}{(c) Teaspoon, Curry} \\
            \includegraphics[height=2.8cm]{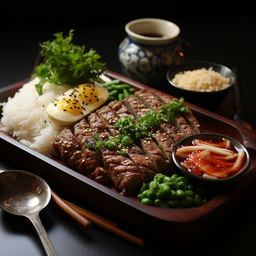} &
            \includegraphics[height=2.8cm]{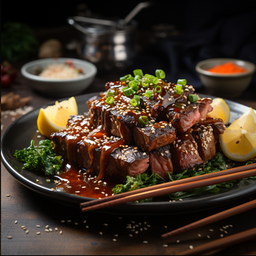} &
            \includegraphics[height=2.8cm]{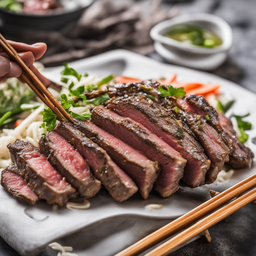} \\
             \multicolumn{3}{c}{(d) Savory steak, Chopsticks} \\
            \includegraphics[height=2.8cm]{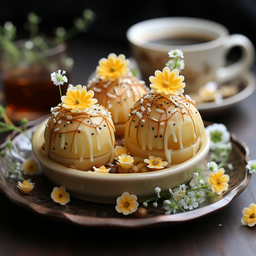} &
            \includegraphics[height=2.8cm]{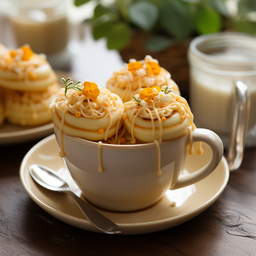} &
            \includegraphics[height=2.8cm]{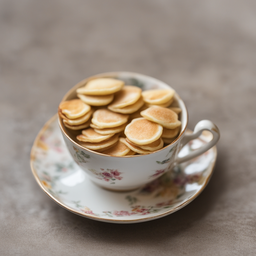} \\
             \multicolumn{3}{c}{(e) Pancakes, Tea cup} \\
            \end{tabular}
        }
    \end{center}
    \caption{Visualizations of images at Level 1 - 4 generated by baseline models and our \dynamic. With additional information, baseline models can indeed generate a small number of correct images, and our \dynamic is able to increase the frequency of correct generation.}
    \label{fig:level1-4_fix_a}
\end{figure*}

\begin{figure*}[h]
    \begin{center}
        \begin{tikzpicture}
            \node[draw=none,fill=none,text=black,font=\fontsize{10}{10}\selectfont] at (0.25\linewidth, 0) {Baseline Models};
            \node[draw=none,fill=none,text=black,font=\fontsize{10}{10}\selectfont] at (0.5\linewidth, 0.4) {Baseline Models};
            \node[draw=none,fill=none,text=black,font=\fontsize{10}{10}\selectfont] at (0.5\linewidth, 0) {with Additional Info};
            \node[draw=none,fill=none,text=black,font=\fontsize{10}{10}\selectfont] at (0.74\linewidth, 0) {\dynamic};
        \end{tikzpicture}
        \adjustbox{max width=\linewidth}{
            \begin{tabular}{c c c}
            \includegraphics[height=2.8cm]{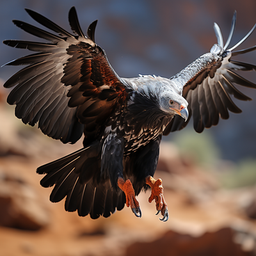} &
            \includegraphics[height=2.8cm]{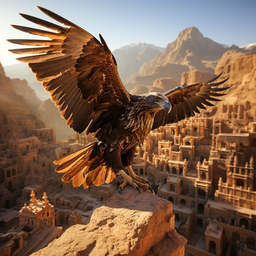} &
            \includegraphics[height=2.8cm]{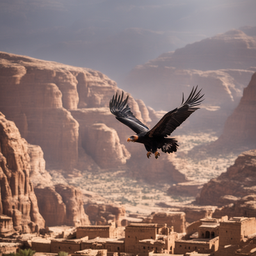} \\
             \multicolumn{3}{c}{(f) California Condor, Petra} \\
            \includegraphics[height=2.8cm]{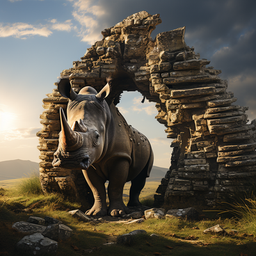} &
            \includegraphics[height=2.8cm]{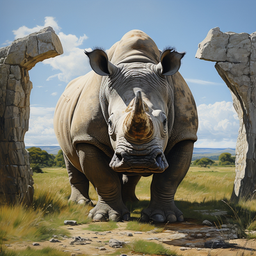} &
            \includegraphics[height=2.8cm]{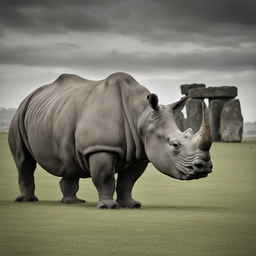} \\
             \multicolumn{3}{c}{(g) Javan Rhino, Stonehenge} \\
            \includegraphics[height=2.8cm]{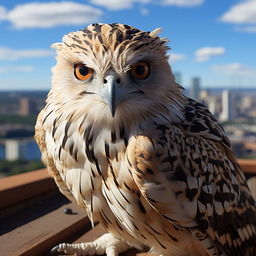} &
            \includegraphics[height=2.8cm]{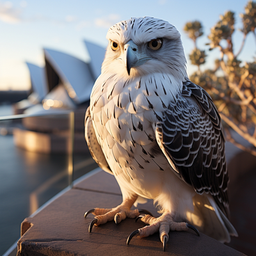} &
            \includegraphics[height=2.8cm]{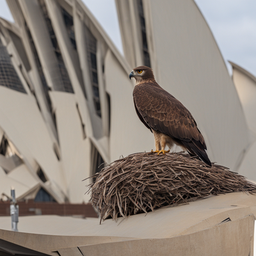} \\
             \multicolumn{3}{c}{(h) Javan Halk-Eagle, Sydney Opera House} \\
            \includegraphics[height=2.8cm]{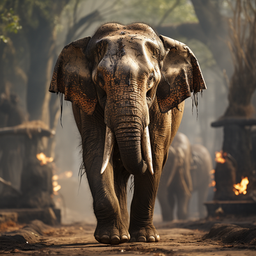} &
            \includegraphics[height=2.8cm]{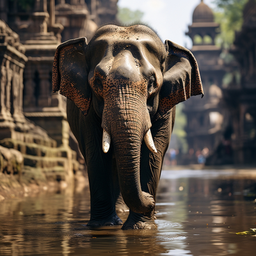} &
            \includegraphics[height=2.8cm]{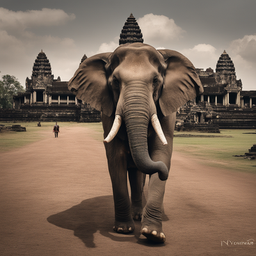} \\
             \multicolumn{3}{c}{(i) Asian Elephant, Angkor Wat} \\
            \includegraphics[height=2.8cm]{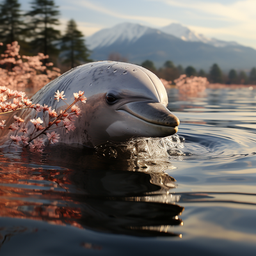} &
            \includegraphics[height=2.8cm]{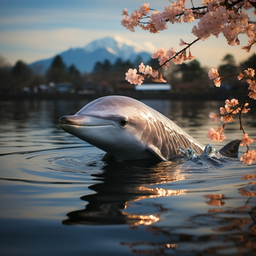} &
            \includegraphics[height=2.8cm]{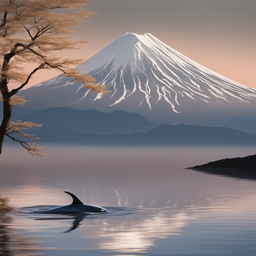} \\
             \multicolumn{3}{c}{(j) Indus River Dolphin, Mount Fuji} \\
            \end{tabular}
        }
    \end{center}
    \caption{Visualizations of images at Level 1 - 4 generated by baseline models and our \dynamic. With additional information, baseline models can indeed generate a small number of correct images, and our \dynamic is able to increase the frequency of correct generation.}
    \label{fig:level1-4_fix_b}
\end{figure*}

For images restored by baseline models by adding rich information, our \dynamic can also easily retrieve the lost concepts and increase the frequency of correct generation, as presented in Figure~\ref{fig:level1-4_fix_a} and~\ref{fig:level1-4_fix_b}.

\end{document}